\documentclass{article}

\usepackage{graphicx}
\usepackage{epstopdf}
\usepackage[T1]{fontenc}
\usepackage{subfigure}
\usepackage{amsmath}
\allowdisplaybreaks[4]
\usepackage{hyperref}
\usepackage{mathrsfs}
\usepackage{amsfonts}
\usepackage{xcolor}
\usepackage[ruled]{algorithm2e}
\usepackage{microtype}
\usepackage{bm}

\usepackage{multirow}
\usepackage{enumerate}
\usepackage{amssymb}
\usepackage{appendix}
\usepackage{makecell}
\usepackage{bbm}
\usepackage{appendix}
\usepackage{pgf}
\usepackage{tikz}
\usepackage{verbatim}
\usetikzlibrary{arrows, decorations.pathmorphing, backgrounds, positioning, fit, petri, automata}

\definecolor{pop}{rgb}{0.6,0.2,1}

\newcommand{\cO}{\mathcal{O}}

\newtheorem{theo}{\textbf{Theorem}}
\newtheorem{def1}{\textbf{Definition}}

\newtheorem{lem}[theo]{\textbf{Lemma}}

\newtheorem{rem}{\textbf{Remark}}[section]
\newtheorem{exa}{\textbf{Example}}[section]
\newtheorem{cla}{\textbf{Claim}}[section]
\newtheorem{pro}{\textbf{Proposition}}[section]
\newcommand*{\QEDA}{\hfill\ensuremath{\blacksquare}}%

\newcommand{\dataset}{{\cal D}}

\begin{document}

\title{On the Provable Generalization of Recurrent Neural Networks}

\author{Lifu Wang, Bo Shen, Bo Hu, Xing Cao\\
	Beijing Jiaotong University\\
	\texttt{\{Lifu\_Wang, bshen, hubo2018,caoxing\}@bjtu.edu.cn}
}

\date{}
\maketitle
\begin{abstract}
Recurrent Neural Network (RNN) is a fundamental structure in deep learning. Recently, some works study the training process of over-parameterized neural networks, and show that over-parameterized networks can learn functions in some notable concept classes with a provable generalization error bound. In this paper, we analyze the training and generalization for RNNs with random initialization, and provide the following improvements over recent works:
\begin{enumerate}[(1)]
\item For a RNN with input sequence $x=(X_1,X_2,...,X_L)$, previous works study to learn functions that are summation of $f(\beta^T_lX_l)$ and require normalized conditions that $||X_l||\leq\epsilon$ with some very small $\epsilon$ depending on the complexity of $f$. In this paper, using detailed analysis about the neural tangent kernel matrix, we prove a generalization error bound to learn such functions without normalized conditions and show that some notable concept classes are learnable with  the numbers of iterations and samples scaling almost-polynomially in the input length $L$.
\item  Moreover, we prove a novel result to learn N-variables functions of input sequence  with the form $f(\beta^T[X_{l_1},...,X_{l_N}])$, which do not belong to the ``additive'' concept class, i,e., the summation of function $f(X_l)$. And we show that when either $N$ or $l_0=\max(l_1,..,l_N)-\min(l_1,..,l_N)$ is small, $f(\beta^T[X_{l_1},...,X_{l_N}])$ will be learnable with the number iterations and samples scaling  almost-polynomially in the input length $L$.

\end{enumerate}
\end{abstract}
\section{Introduction}
In Deep Learning, the recurrent neural network (RNN) is well-known as one of the most popular models to model sequential data and is widely used in practice for tasks in natural language processing (NLP). One of the characters of RNN is that it performs the same operation for all the input of the sequence.

Consider a input sequence $x=(X_1,X_2,...,X_L)$. A RNN with the form
\begin{equation}\label{rnne}
h_{l}(x)=\phi(\bm{W}h_{l-1}+\bm{A}X_l),
\end{equation}
 is trying to learn functions $f_l(X_1, X_2,...X_l)$ as
\begin{equation}
\begin{aligned}
&h_{1}(x)=f_1(X_1)\\
&h_{2}(x)=f_2(X_1, X_2)\\
&\ \ \ \ \ \ \ \ \ \ \ \ \ \ \vdots\\
&h_{L}(x)=f_L(X_1,X_2,...X_L)
\end{aligned}
\end{equation}
Due to the complex nonlinearity, the loss is generally non-convex, and it is very difficult to give a theoretical guarantee. Recently, there are some works \cite{allenzhu2019convergence,cao,NEURIPS201962dad6e2,DBLP,sa,rate} trying to give a theoretical explanation that why gradient descent can allow an overparametrized  network to attain arbitrarily low training error and ample generalization ability.  These papers show that, under some assumptions, we have:
\begin{itemize}
\item[$\bullet$] \emph{Multi-layer feed-forward networks \cite{allenzhu2019convergence,DBLP} and recurrent neural networks \cite{rate} with large hidden size can attain zero training error, regardless of whether the data is properly labeled or randomly labeled.}
\item[$\bullet$] \emph{For multi-layer feed-forward networks, functions with the form $F^*(x)=\sum_{r=1}^{C} \phi_{r}(\beta^T_{r}X), X\in\mathbb{R}^d,\beta_r\in\mathbb{R}^d, ||\beta_r||=1$
     are learnable i.e. fitting the training data with a provably small generalization error, if $\phi$ is analytic  and the ``complexity'' is low enough \cite{NEURIPS201962dad6e2,sa,cao}.}
\item[$\bullet$] \emph{The ``complexity'' of function $\phi$ can be measured by a matrix derived from the NTK (Neural Tangent Kernel) of the network \cite{sa,cao}.}
\item[$\bullet$] \emph{For recurrent neural networks \cite{allenzhu2019sgd}, if the input sequence is normalized, i.e., $x=(X_1,X_2,...,X_L)$, $||X_1||=1$, $||X_l||=\epsilon$ with $\epsilon$ very small,  functions with the form $F^*(x)=\sum_{l=1}^L \sum_{r=1}^{C_l} \phi_{l,r}(\beta^T_{l,r}X_l)$ are learnable, where $m$ is the size of matrix $\bm{W}$, and $\mathscr{C}=\sum_{i=0}^\infty a_iR^i $ is a series representing the complexity of learnable functions.}
\end{itemize}
These works show the provable learning ability of deep learning.
But there are still some important issues that were not addressed.
\begin{itemize}
	\item[$\bullet$] Firstly, for RNNs, the method in \cite{allenzhu2019sgd} requires a normalized condition for $\bm{A}$ and $X_l$ in (\ref{rnne}) that $||\bm{A}X_l||\leq \epsilon_x$ for all $l\leq L$ and shows that for a function $F^*(x)$ with the complexity $\mathscr{C}$, it is learnable with error $O(\epsilon^{1/3}_x\mathscr{C})$.  Thus $ ||X_l||$ (or equally, $||\bm{A}||$) should be very small and the scale is dependent on the complexity of functions. The dependence of $||\bm{A}X_l||$ on $\mathscr{C}$ makes the results unrealistic in practice since generally the norm of input will not be so small.
	
	\item[$\bullet$]Secondly, the result in  \cite{allenzhu2019sgd} shows that RNNs can learn functions which are the summation of functions like $\psi(\beta^T_{l}X_l)$. But this is only a linear combination of the functions of the input at different positions and does not consider the nonlinear interaction of the inputs. One may ask, since $h_{L}(x)$ is a function of $\{X_1,X_2,...X_L\}$, is it possible to go beyond and learn more complex functions?
\end{itemize}

In order to study these problems, we consider the binary classification  problem: for every input $x_i$, the label ($+1$ or $-1$) of $x_i$ can be expressed by the sign of a target function  $F^*(x_i)$.  We consider Elman recurrent neural networks with ReLU activation
\begin{equation}\label{rnn}
\begin{aligned}
&h_{l}(x)=\phi(\bm{W}h_{l-1}+\bm{A}X_l)\\
&f(\bm{W},x)=\bm{B}^Th_L(x)\in\mathbb{R}.\\
&x=(X_1,X_2,...,X_L),X_l\in\mathbb{R}^d,\bm{W}\in\mathbb{R}^{m\times m},\\
&\bm{A}\in\mathbb{R}^{m\times d},\bm{B}\in\mathbb{R}^m,\phi(x)=\max (x,0)
\end{aligned}
\end{equation} to learn two types of target functions:
\begin{itemize}
\item[$\bullet$]{ Additive Concept Class:}
\begin{equation}\label{cc1}
\begin{aligned}
&F^*(x)=\sum_{l=1}^{L} \sum_{r=1} \psi_{l,r}(\beta^T_{l,r}X_l/||X_l||),\\
&\psi_{l,r}(x)=\sum_{i=0}^\infty c_ix^{i},\\
\end{aligned}
\end{equation}
\item[$\bullet$]{N-variables Concept Class:}
\begin{equation}\label{cc2}
\begin{aligned}
&F^*(x)=\sum_{r}\psi_{r}(\langle \beta_{r}, [X_{l_1},...,X_{l_{N}}]\rangle),\\
&\psi_{r}(x)=\sum_{i=0}^\infty c_ix^{i}.
\end{aligned}
\end{equation}
\end{itemize}
For these two types of function,  we study the following  questions:
\begin{itemize}
\item[$\bullet$] Can RNN learn additive concept class functions (\ref{cc1}) without the normalized condition with {\bf reasonable complexity on the sequence size $L$}?
\item[$\bullet$] Can RNN learn functions in N-variables Concept Class (\ref{cc2}) which can not be written as the summation of $f(X_l)$ with {\bf reasonable complexity on  $N$ and $L$}?
\end{itemize}
{\bf Our Result. }
We answer the two questions and give a provable generalization error bound. Our results are stated as follows:
\begin{theo}(Informal) For a function $F^*(X_1,X_2,...,X_L)$ with the form as in (\ref{cc1}) or (\ref{cc2}),  there is a power series named the complexity $\mathscr{C}(F^*)$ dependent on the Taylor expansion coefficient in  (\ref{cc1}) and (\ref{cc2}). For (\ref{cc1}), $\mathscr{C}(F^*)$  is almost-polynomial in $L$. For (\ref{cc2}), when $N$ or $l_0=\max(l_1,..,l_N)-\min(l_1,..,l_N)$ is small, $\mathscr{C}(F^*)$ is almost-polynomial in $L$. Under this definition of complexity $\mathscr{C}(F^*)$, $F^*$ is learnable using RNN with $m$ hidden nodes and ReLU activation in (\ref{rnn}) in $\cO(\mathscr{C}(F^*)^2)$ steps with $\cO(\mathscr{C}(F^*)^2)$ samples if $m\geq poly(L,\mathscr{C}(F^*))$.
\end{theo}

{\bf Contribution.} We summarize the contributions as follows:
\begin{itemize}
\item[$\bullet$] \emph{In this paper, we prove that RNN without normalized condition can efficiently learn some notable concept classes  with {\bf both time and sample complexity scaling almost polynomially in the input length $L$}.}
\item[$\bullet$] \emph{Our results go beyond the ``additive'' concept class. We prove a novel result that RNN can learn more complex function of the input such as N-variables concept class functions. And ``long range correlation functions'' with small $N$ (e.g. $N=2$, $f(\beta^T[X_l,X_{l+l_0}])$ ) are learnable with {\bf complexity scaling almost polynomially in the input length $L$ and correlation distance $l_0$}.}
\item[$\bullet$] \emph{Technically, we study the ``backward correlation'' of RNN network. In RNN case, using a crucial observation on the degeneracy of deep network, we show that the {\bf ``backward correlation'' $\frac{1}{m}\langle \text{Back}_l(x_i),\text{Back}_{l}(x_j) \rangle$ will decay polynomially rather than exponentially in input length $L$}. This shows the complexity of learning RNN with ReLU activation function is polynomial in the size of input sequence $L$.}
\end{itemize}

{\bf Notions.}
For two matrices $\bm{A}, \bm{B}\in \mathbb{R}^{m\times n}$, we define $\langle \bm{A}, \bm{B} \rangle= \text{Tr}(A^TB)$. We define the asymptotic
notations $\cO(\cdot), \Omega(\cdot),poly(\cdot)$ as follows.  $a_n, b_n$ are two sequences. $a_n=\cO(b_n)$ if $\lim \sup_{n\to \infty}|a_n/b_n|< \infty$, $a_n=\Omega(b_n)$ if
$\lim \inf_{n\to \infty}|a_n/b_n|>0$, $a_n=poly(b_n)$ if there is $k\in \mathbb{N}$ that $a_n=O((b_n)^k)$. $\widetilde{\cO}(\cdot), \widetilde{\Omega}(\cdot), \widetilde{poly}(\cdot)$ are notions which hide the logarithmic factors in $\cO(\cdot), \Omega(\cdot), poly(\cdot)$. $||\cdot ||$ and $||\cdot||_2$ denote the 2-norm of matrices. $||\cdot ||_1$ denote the 1-norm. $||\cdot ||_F$ is the Frobenius-norm. $||\cdot||_0$ is the number of non-zero entries.

For elements $A_{i.j}, B_{i,j}$ of symmetric matrix $\bm{A}, \bm{B}$. We abuse the notion $A_{i.j}\succeq B_{i.j}$ to denote $\bm{A}\succeq \bm{B}$, i.e. $\bm{A}-\bm{B}$ is a positive semidefinite matrix.
\section{Preliminaries}
\subsection{Function Complexity}\label{fc}
For a analytic  function $\psi(z)$, we can write it as $\psi(z)=  c_0+ \sum_{i=1}^\infty c_{i}z^{i}$. We define the following notion to measure the complexity to learn such functions.
\begin{equation}
\mathscr{C}(\psi,R)= 1+\sum_{i=1}^\infty i\cdot |c_{i}|R^{i}.
\end{equation}
\begin{equation}
\mathscr{C}_N(\psi, R)= 1+\sum_{i=1}^\infty  L^{1.5N} C_1^N\cdot \sqrt{C_{N,i}}\cdot (i/N)^N  \cdot |c_{i}|R^{i}
\end{equation}
where $C_1>100$ is an large absolute constant and $C_{N,i}$ is the  largest combination number $\frac{i!}{n_1!n_2!...n_N!}$ for $n_1,n_2...n_N>0, n_1+n_2+...n_N=i$,
\begin{exa} \cite{sa}
Consider $\psi(z)= arctan(z/2)$. Then
\begin{equation}
\psi(z)=\sum_{i=1}\frac{(-1)^{i-1}2^{1-2i}}{2i-1}z^{2i-1}
\end{equation}
In this case, $$\mathscr{C}(\psi,1)=1+\sum_{i=1}^\infty i\cdot |c_{i}|\leq 1+\sum_{i=1}^\infty 2^{1-2i}\leq \cO(1).$$
\end{exa}
\begin{exa}
In the case $N=2$, $C_{2,i}=i$, $(i/2)^2\leq i^2$. $\psi(z)= exp(z)$
$$\mathscr{C}_2(\psi,1)\leq1+\sum_{i=1}^\infty L^3C_1^2 \pi i^{2.5}/i!\leq \cO(1)$$
\end{exa}

\subsection{Concept Class}\label{fcc}
For the input sequence $\{X_l\}$, we assume $C_{min}\leq ||X_l||\leq C_{max}$, for all $1\leq l\leq L$ and $C_{max}/C_{min}\sim C_0$. Under this condition, we consider two types of target functions with the following form:\\
{\bf Additive Concept Class.}
\begin{equation}\label{cop}
F^*(x)=\sum_{l=1}^{L} \sum_{r=1}^{C_l} \psi_{l,r}(\beta^T_{l,r}X_l/||X_l||).
\end{equation}
Here for all $l,r$, $\psi_{l,r}$ is analytic  and $||\beta_{l,r}||_2\leq 1$.

We define
\begin{equation}\label{com1}
\mathscr{C}(F^*)=L^{3.5}\sum_{l=1}^L \sum_{r=1}^{C_l} \mathscr{C}(\psi_{l,r},C_0\sqrt{ L}),
\end{equation}
to be the complexity of the target function.
\begin{rem}
If we consider function $\psi(\beta^TX_l)$ and $||X_l||=1$ for all $l$, the above complexity will become  $\mathscr{C}(\psi,\cO(\sqrt{L}))$. This is similar with that in \cite{allenzhu2019sgd} but this complexity requirement is much weaker than that in \cite{allenzhu2019sgd}. For example, the complexity of $arctan(z/2)$ in \cite{allenzhu2019sgd} is not finite, as shown in \cite{sa}.
\end{rem}
{\bf N-variables Concept Class.}
\begin{equation}\label{cop2}
F^*(x)=\sum_{r}\psi_{r}(\langle \beta_{r}, [X_{l_1},...,X_{l_{N}}]\rangle/\sqrt{N}\max ||X_{l_n}||).
\end{equation}
For all $r$, $\psi_{l,a,r}(x,y)$ is an analytic  function $\psi_{r}(x)=c_0+ \sum_{i=1}^\infty c_ix^{i}$. $\beta_{r}\in \mathbb{R}^{dN}$, $||\beta_{r}||_2\leq 1$. Let $l_0=\max(l_1,..,l_N)-\min(l_1,..,l_N)$.
We define
\begin{equation}\label{com2}
\mathscr{C}(F^*)=\min(L^{2}\mathscr{C}_N(\psi_{r},C_0\sqrt{ L}), L^{3.5}\mathscr{C}(\psi_{r},2^{l_0}C_0\sqrt{ L})).
\end{equation}
\begin{rem}
The complexity $\sum_{r} \mathscr{C}_N(\psi_{r},C_0\sqrt{ L})$ and $\sum_{r}\mathscr{C}(\psi_{r},2^{l_0}C_0\sqrt{ L})$ are exponential in $N$ and $l_0$ respectively. And $\mathscr{C}(F^*)$ is less or equal than both. Thus if either $l_0$ or $N$ is small, $\mathscr{C}(F^*)$ will be polynomial in $L$. Especially when $N$ is small(e.g. N=2), even if $l_0=L-1$, functions with the form $f(\beta^T[X_l,X_{l+l_0}])$ are still learnable with a low complexity.
\end{rem}

\subsection{Results on Positive Definite Matrices and Functions}
We say a function $\phi(\cdot,\cdot):\mathbb{R}^d\times \mathbb{R}^d\to \mathbb{R}$ is positive definite if for all $n \in \mathbb{N}$, any $\{x_1,...,x_n\}\subseteq \mathbb{R}^d, \{c_1,...,c_n\}\subseteq \mathbb{R}$,
\begin{equation}
\sum_{i,j}c_ic_j\phi(x_i,x_j)\geq 0.
\end{equation}

The following basic properties in chapter 3 of \cite{Christian} are very useful in our proof.
\begin{pro}
If $\phi(\cdot,\cdot)$ is positive definite function, let matrix $\bm{M}\in \mathbb{R}^{n\times n}$, $\{x_1,...,x_n\}\subseteq \mathbb{R}^d$, and $M_{i,j}= \phi(x_i,x_j)$. Then $\bm{M}$ is a semi-positive definite matrix.
\end{pro}
\begin{pro}\label{c1}
If $\phi_1(\cdot,\cdot)$ and $\phi_1(\cdot,\cdot)$ are positive definite, $\phi(x_i,x_j)=\phi_1(x_i,x_j)\cdot \phi_2(x_i,x_j)$ is also a positive definite function.
\end{pro}

\begin{pro}\label{c2}
Let $\phi(\cdot, \cdot)$ be a positive definite function, and $\psi(x)=\sum_{i=0}^\infty c_ix^i$, $c_i\geq 0$. Then $\psi(\phi(\cdot,\cdot))$ is also a positive definite function.
\end{pro}

For a positive definite matrix $\bm{M}\in \mathbb{R}^{n\times n}$, there is a result in  \cite{sa},
\begin{pro}\label{claa}(Section E of \cite{sa}.)
Let $\bm {X}=(x_1,...x_n)\in \mathbb{R}^{d\times n}$ and $\bm{K}_{p}\in \mathbb{R}^{n\times n}$ is a matrix with $(K_{p})_{i,j}=(x_i^Tx_j)^{p}$. Suppose there is $\alpha>0$, such that $\bm{M}\succeq \alpha^2 \bm{K}_{p}$. Let $y=((\beta^Tx_1)^{p},...,(\beta^Tx_n)^{p})\in \mathbb{R}^{n}$. We have $\sqrt{y^T(\bm{M})^{-1}y}\leq ||\beta||_2^{p}/\alpha$.
\end{pro}

\section{Main Results}
Assume there is an unknown data set $\dataset=\{x,y\}$. The inputs have the form $x=(X_1,X_2,...X_L)\in (\mathbb{R}^d)^L$. $||X_l||\leq \cO(1)$ for all $1\leq l\leq L$.  For every input $x_i$, there is a label $y_i=\pm1$.

The neural network with input $x$ is
\begin{equation}\label{ini}
\begin{aligned}
&h_0(x)=\phi(\bm{M}_0),\\
&h_{l}(x)=\phi(\bm{W}h_{l-1}+\bm{A}X_l),\\
&f(\bm{W},x)=\bm{B}^Th_L(x).
\end{aligned}
\end{equation}
Here $ \bm{W}\in\mathbb{R}^{m\times m}, \bm{A}\in \mathbb{R}^{m\times d}, \bm{B}, \bm{M}_0 \in \mathbb{R}^{ m}$. The entries of $\bm{M}_0$, $\bm{W}$ and $\bm{A}$ are respectively i.i.d. generated from $N(0,\frac{2}{m})$, $N(0,\frac{2}{m})$ and $N(0,\frac{2}{L^3\cdot m})$. The entries of $\bm{B}$ are i.i.d. generated from $N(0,\frac{1}{m})$.

The goal of learning RNN is to minimize the population loss:
\begin{equation}\label{ris}
L_{ \dataset}(\bm{W})=\mathbb{E}_{(x,y)\sim \dataset} \ell(y\cdot f(\bm{W},x)),
\end{equation}
by optimizing the empirical loss
\begin{equation}
L_{ S}(\bm{W})=\frac{1}{n}\sum_{i=1}^n\ell(y_i\cdot f(\bm{W},x_i)),
\end{equation}
using SGD.
Here $\ell(x)=\log(1+exp(-x))$ is the cross-entropy loss. Consider the SGD algorithm on this RNN.
\begin{algorithm}\label{a1}
\caption{Training RNN with SGD}
{\bf Input:} Data set $\dataset$, learning rate $\eta$.\\
The entries of $\bm{W}^0,\bm{A}$ are i.i.d. generated from $N(0,\frac{2}{m})$. The entries of $\bm{B}$ are i.i.d. generated from $N(0,\frac{1}{m})$.\\
\For {$t=1,2,3...n$}{
Randomly sample $(x_t, y_t)$ from the data set $\dataset$.\\
$\bm{W}^t=\bm{W}^{t-1}-\eta\nabla_{\bm{W}^{t-1}}\ell(y_t\cdot f(\bm{W}^{t-1},x_t))$.
}
\end{algorithm}
Let the complexity $\mathscr{C}^*$ of $F^*(\cdot)$ be defined in (\ref{com1}) and (\ref{com2}). The 0-1 error for $\dataset$ is $ L^{0-1}_{\dataset}(\bm{W})=\mathbb{E}_{(x,y)\sim \dataset}\mathbbm{1}\{y\cdot f(\bm{W},x)<0\}$. We have:
\begin{theo}\label{tt}
Assume there is $\delta\in (0,e^{-1}]$.  Supposing for $\dataset=\{x_i,y_i\}$, there is a function $F^*$ belonging to the concept class (\ref{cop}) or (\ref{cop2}) such that $y_i\cdot F^*(x_i)\geq 1$ for all $i$.  Let $\bm{W}^k$ be the output of Algorithm \ref{a1}. There is a parameter $m^*(n,\delta,L,\mathscr{C}^*)=poly(n,\delta^{-1},L,\mathscr{C}^*)$  such that, with probability at least $1-\delta$, if $m>m^*(n,\delta,L)$, there exits parameter $\eta= \cO(1/ m)$ that satisfies
\begin{equation}
\frac{1}{n}\sum_{k=1}^nL^{0-1}_{\dataset}(\bm{W}^k)\leq \widetilde{\cO}[\frac{(\mathscr{C}^*)^2}{n}]+\cO(\frac{\log(1/\delta)}{n}).
\end{equation}
\end{theo}

\begin{rem}
This theorem induces that, to achieve {\bf population $0-1$ error}(rather than empirical loss) being less than $\epsilon$, it is enough to train the network using  Algorithm \ref{a1} with $\widetilde{\Omega}((L\cdot\mathscr{C}^*)^2/\epsilon)$ steps. As defined in section \ref{fc} and \ref{fcc}, when $N$ is small, for the two types of concept class, $(\mathscr{C}^*)^2$ is almost-polynomial in input length $L$. Thus they can be learned effectively.
\end{rem}

\begin{rem}
This theorem can also be generalized to ``sequence labeling'' loss such as $\frac{1}{n}\sum_{i=1}^n\sum_{l=1}^L\ell(y_i\cdot f_l(\bm{W},x_i))$ with $f_l(\bm{W},x)=\bm{B}^Th_l(x)$. This is because the matrix $$H^l_{i,j}=\frac{1}{m}\langle \nabla f_l(\bm{W},x_i), \nabla f_l(\bm{W},x_j)\rangle$$ with different $l$ are almost ``orthogonal'' by a similar argument to (\ref{ot}) in Theorem \ref{tt6}. Then RNN can learn a function $f_l=sign(F^*_l(x))$ with $F^*_l(x)$ belonging to functions in section \ref{fcc}. See Remark \ref{g1} in the supplementary materials.
\end{rem}

\section{Sketch Proof of the Main Theorem}
The first step to prove the main theorem \ref{tt} is the following generalization of Corollary 3.10 in \cite{cao}.
\begin{theo}\label{mt}
Under the condition of Theorem \ref{tt}, let $n$ samples in the training set be $\{x_i,y_i\}_{i=1}^n$. $\widetilde{y}=[F^*(x_1),F^*(x_2),...F^*(x_n)]^T$. Let $\bm{H}$ be a matrix with $H_{i,j}= \frac{1}{m}\langle \nabla_{\widetilde{W}} f(\widetilde{\bm{W}},x_i),\nabla_{\widetilde{W}} f(\widetilde{\bm{W}},x_j) \rangle$. The entries of $\widetilde{\bm{W}}$ are i.i.d. generated from $N(0,\frac{2}{m})$.
If there is a matrix $\bm{H}^\infty \in \mathbb{R}^{n\times n}$ satisfying
\begin{equation}\label{condition1}
\bm{H} +\bm{\epsilon}^T\bm{\epsilon} \succeq \bm{H}^\infty \text{ with } ||\bm{\epsilon}||_F\leq 0.01/\cO(\mathscr{C}^*),
\end{equation}
and $\sqrt{\widetilde{y}^T (\bm{H}^\infty)^{-1}\widetilde{y}}\leq \cO(\mathscr{C}^*)$,
there exits $m^*(n,\delta^{-1},L,\mathscr{C}^*)=poly(n,\delta^{-1},L,\mathscr{C}^*)$ such that, with probability at least $1-\delta$, if $m>m^*$,
\begin{equation}
\frac{1}{n} \sum_{k=1}^n L^{0-1}_{\dataset}(\bm{W}^k)\leq \widetilde{\cO}[\frac{\widetilde{y}^T (\bm{H}^\infty)^{-1}\widetilde{y}}{n}]+\cO(\frac{\log(1/\delta)}{n}).
\end{equation}
\end{theo}
\begin{rem}
In order to show Theorem \ref{tt} using this theorem, we need to carefully pick out the exponential parts of $L$. Using the methods in \cite{rate} and \cite{cao}, we can show that $m^*(L,n,\sqrt{\widetilde{y}^T (\bm{H}^\infty)^{-1}\widetilde{y}})\geq poly(n,L,\sqrt{\widetilde{y}^T (\bm{H}^\infty)^{-1}\widetilde{y}})$ is enough.   $\sqrt{\widetilde{y}^T (\bm{H}^\infty)^{-1}\widetilde{y}}$ is dealt with by calculating the forward and backward correlation in section \ref{f1} and \ref{f2}.
\end{rem}
The proof of theorem \ref{mt} is in fact a combination of the results in \cite{cao} and \cite{rate}. The really matter thing is how large can $\sqrt{\widetilde{y}^T (\bm{H}^\infty)^{-1}\widetilde{y}} \ $ be. We can show that:
\begin{theo}\label{mc}
Under the condition of Theorem \ref{mt}, with probability at least $1-\delta$, there exits matrix $\bm{H}^\infty$ satisfying (\ref{condition1}) and
\begin{equation}\label{ec}
\sqrt{\widetilde{y}^T (\bm{H}^\infty)^{-1}\widetilde{y}}\leq \cO(\mathscr{C}^*).
\end{equation}
\end{theo}
Theorem \ref{tt} is a direct corollary of the above two theorems.
\subsection{Calculation on Kernel Matrix}
The proof of (\ref{ec}) relies on a direct calculation to construct a kernel matrix $\bm{H}^\infty$.  We consider two input $x_i$ and $x_j$. Let $X_{i,l}$ and $X_{j,l}$ be the $l-th$ input of $x_i$ and $x_j$. Let $D_l\in \mathbb{R}^{m\times m}$ and $D_l'\in \mathbb{R}^{m\times m}$ be diagonal matrices that,
\begin{equation}
\begin{aligned}
&(D_l)_{k,k}=\mathbbm{1}\{\bm{W}h_{l-1}(x_i)+\bm{A}X_{i,l}>0\}\\
&(D'_l)_{k,k}=\mathbbm{1}\{\bm{W}h_{l-1}(x_j)+\bm{A}X_{j,l}>0\}
\end{aligned}
\end{equation}

\begin{equation}
\begin{aligned}
\text{Back}_l=BD_LW\cdots D_{l+1}W, \text{Back}'_l=BD'_LW\cdots D'_{l+1}W
\end{aligned}
\end{equation}

Then
\begin{equation}
\frac{1}{m}\langle \nabla_{\widetilde{W}} f(\widetilde{\bm{W}},x_i),\nabla_{\widetilde{W}} f(\widetilde{\bm{W}},x_j) \rangle=\frac{1}{m}\sum_{l,l'}\langle \text{Back}_l(x_i)\cdot D_l,\text{Back}_{l'}(x_j)\cdot D_{l'}' \rangle\cdot \langle h_l(x_i), h_{l'}(x_j)\rangle
\end{equation}
 Generally $H_{i,j}= \frac{1}{m}\langle \nabla_{\widetilde{W}} f(\widetilde{\bm{W}},x_i),\nabla_{\widetilde{W}} f(\widetilde{\bm{W}},x_j) \rangle$ is hard to deal with. However, in the $m\to \infty$ limit, we can use some techniques to do the calculation.

\subsubsection{Forward Correlation}\label{f1}
\begin{theo}\label{tt5}
For fixed $i,j$, under the condition in Theorem \ref{mt}, with probability at least $1-exp(-\Omega(\log^2m))$,
\begin{equation}
|\langle h_l(x_i),h_l(x_j) \rangle -K^l_{i,j}|\leq \cO(l^{16}\cdot\log^2 m/\sqrt{m})
\end{equation}
And let $Q_l=\sqrt{(1+\frac{1}{L^3}\sum_{k=1}^l||X_{i,k}||^2)\cdot(1+\frac{1}{L^3}\sum_{k=1}^l||X_{j,k}||^2) }$,
\begin{equation}\label{ek}
\begin{aligned}
&K^1_{i,j}= Q_1\cdot \sum_{r=0}^\infty \mu^2_r [(1+\frac{1}{L^3}X_{i,1}^TX_{j,1})/Q_1]^r\\
&K^{l}_{i,j}=Q_l\cdot \sum_{r=0}^\infty \mu^2_r (\{\frac{1}{L^3}X_{i,l}^TX_{j,l}+ K^{l-1}_{i,j}\}/Q_l)^r
\end{aligned}
\end{equation}
In the above equations, $\mu_r=\frac{1}{\sqrt{2\pi}}\int_0^\infty \sqrt{2}xh_r(x)e^{-\frac{x^2}{2}}dx$, $h_r(x)=\frac{1}{\sqrt{r!}}(-1)^re^{\frac{x^2}{2}}\frac{d^r}{dx^r}e^{-\frac{x^2}{2}}.$
\end{theo}
\subsubsection{Backward Correlation}\label{f2}
\begin{theo}\label{tt6} For $l\neq l'$, with probability at least $1-exp(-\Omega(\log^2m))$,
\begin{equation}\label{ot}
|\frac{1}{m}\langle \text{Back}_l(x_i)\cdot D_l,\text{Back}_{l'}(x_j)\cdot D_{l'}' \rangle| \leq \cO(\frac{L^4\log^4 m}{m^{1/4}}).
\end{equation}
For $l=l'$, there is $F^l_{i,j}$ that, with probability at least $1-exp(-\Omega(\log^2m))$,
\begin{equation}
|\frac{1}{m}\langle \text{Back}_l(x_i)\cdot D_l,\text{Back}_{l}(x_j)\cdot D_{l'}' \rangle- F^l_{i,j}|\leq \cO(\frac{L^4\log^4 m}{m^{1/4}}).
\end{equation}
where
\begin{equation}
\Sigma(x)=\frac{1}{2}+\frac{arcsin(x)}{\pi},
\end{equation}
\begin{equation}\label{ef}
F^l_{i,j}\succeq \frac{1}{K}\Sigma( \{\frac{1}{L^3}\langle X_{i,l}, X_{j,l} \rangle +K^{l-1}_{i,j}\}/Q_l).
\end{equation}
and $0<K\leq \cO(1/L^4).$
\end{theo}
\begin{rem}
We should note that this theorem is one of the key differences between this work and the methods in \cite{allenzhu2019sgd}. In fact, we must show that there is a constant $K>0$ such that $\frac{1}{m}\langle \text{Back}_l(x_i),\text{Back}_{l}(x_j) \rangle -K $ is still positive definite. However, is $K$ large enough thus $1/K\geq poly(L)$ rather than  $1/K \leq exp(-\Omega(L))$ ? This is not a trivial question. One can only get $K \geq \frac{1}{2^L}$ using naive estimation. In \cite{allenzhu2019sgd},  $||\bm{A}X_l||\leq \epsilon_x$ is required to  make sure $\text{Back}'_l=\text{Back}_l(x_i)-\text{Back}_l(x_j)$ samll. However after $k$ steps of training, we can show the approximation error is roughly $\cO(||\text{Back}'||\cdot ||\bm{W}^k-\bm{W}^0||)$ and $||\bm{W}^k-\bm{W}^0||_F \sim \sqrt{\widetilde{y}^T(\bm{H}^\infty)^{-1}\widetilde{y}} \sim \mathscr{C}(F^*)$. Thus the dependence of $\epsilon_x $ on  $\mathscr{C}(F^*)$ is hard to be dealt with using this method. In this paper, we do not need the normalized condition. Our methods rely on a crucial observation that the function $\lim_{l\to \infty} h_l(x_i)^Th_l(x_j)/(||h_l(x_i)||\cdot||h_l(x_j||)$ will degenerate to a constant function.
\end{rem}

\subsubsection{Sketch Proof of Theorem \ref{mc}}
In order to estimate the complexity, we use the results in the last subsection and Proposition  \ref{claa},\ref{c1} and \ref{c2}.

Proposition \ref{claa} shows that, in order to estimate $\sqrt{\widetilde{y}^T (\bm{H}^{\infty})^{-1}\widetilde{y}}$, we need to show
\begin{equation}
\bm{H}^\infty \succeq \xi_p\cdot  (\bm{X}_l^T\bm{X}_l)^{\circ p}
\end{equation}
with $\xi_p>0$ for all $p\in \mathbb{N}, 1\leq l\leq L$.
Here $\bm{X}_l\in \mathbb{R}^{n\times d}=[X_{1,l}, X_{2,l}... X_{n,l}]$ and
\begin{equation}
[(\bm{X}_l^T\bm{X}_l)^{\circ p}]_{i,j}=\{X_{i,l}^TX_{j,l}\}^p.
\end{equation}
 We will show that, there is a matrix $H^{\infty}$. With probability at least $1-\delta$, $H_{ij}=H^{\infty}_{ij}\pm  \cO(\frac{L^4\log^4 m}{m^{1/4}})$ for all $i,j\in [n]$, and,
\begin{equation}\label{eh}
H^{\infty}_{i,j}\succeq\frac{1}{\cO(L^4)}\cdot Q_l\Sigma(\{\frac{1}{L^3}\langle X_{i,l}, X_{j,l} \rangle + K^{l-1}_{i,j}\}/Q_l).
\end{equation}
for all $l$.

Based on (\ref{eh}), we can show  the following results:

 For all $1\leq l\leq L$ and all $k$
\begin{equation}
H^{\infty}_{i,j}\succeq \frac{1}{\cO(L^4)} \Sigma(\{K^{l}_{i,j}+\frac{1}{L^3}X_{i,l}^TX_{j,l}\}/Q_{l})\succeq  \Omega( \frac{1}{L^7}) \cdot (\frac{1}{\cO(L)})^{k}\cdot \frac{1}{k^2} (X_{i,l}^TX_{j,l})^{k}/(||X_{i,l}||\cdot X_{j,l}||)^{k}.
\end{equation}
This deduces the complexity for the Additive Concept Class in section \ref{fc},
\begin{equation}
\sqrt{\widetilde{y}^T (\bm{H}^{\infty})^{-1}\widetilde{y}}\leq \cO(\mathscr{C}^*).
\end{equation}

As for N-Variables Concept Class,
\begin{equation}
\begin{aligned}
H^\infty_{i,j}\succeq&  \frac{1}{C_1^N L^4 \cdot L^{2N} \cdot C_{N,p}\cdot(p/N)^N}\\
&\cdot (X_{i,r_1}^TX_{j,r_1} +X_{i,r_2}^TX_{j,r_2}... + X_{i,r_N}^TX_{j,r_N})^p/(N\cdot \max_n(||X_{i,r_n}||)\cdot \max_n(||X_{j,r_n}||)) ^{p}
\end{aligned}
\end{equation}
with some large constant $C_1>0$. Meanwhile, for any $l\leq L, a<l$, let $Z_{i, l,a}= [X_{i,l}, X_{i,l-1},... X_{i,l-a}] $. We have:
\begin{equation}
H^\infty_{i,j}\succeq \Omega( \frac{1}{L^7}) \cdot (\frac{1}{\cO(L)})^{k}\cdot \frac{1}{k^2} (Z_{i,l,a}^TZ_{j,l,a})^{k}/(||Z_{i,l,a}||\cdot Z_{j,l,a}||\cdot 2^a)^{k}
\end{equation}

Then from definition of complexity in section \ref{fcc} and Proposition \ref{claa}, we can prove
\begin{equation}
\sqrt{\widetilde{y}^T (\bm{H}^{\infty})^{-1}\widetilde{y}}\leq \cO(\mathscr{C}^*).
\end{equation}

Therefore (\ref{ec}) follows.
\section{Dissicusion}
In this paper, we use a new method to avoid the normalized conditions. The main idea is to provide an esitmation for $\sqrt{\widetilde{y}^T (\bm{H}^\infty)^{-1}\widetilde{y}}$ in the RNN case directly. However, the value of $\sqrt{\widetilde{y}^T (\bm{H}^\infty)^{-1}\widetilde{y}}$ is only explicitly calculated for the two-layer case in \cite{sa}. In the RNN cases, the neural tangent kernel matrix involves the depth and the weight sharing in the network and difficult to deal with.

In   \cite{allenzhu2019sgd}, their method is to reduce the RNN case to $$f_L\approx\sum_lBack^{(0)} \cdot \mathbbm{1}_{\langle W, h_{l-1}\rangle+AX_l\geq 0} W^* \cdot h_{l-1},$$
which is similar to a summation of $L$ two-layer networks. And this reduction requires the following  operations in  \cite{allenzhu2019sgd}:
\begin{enumerate}[1)]
	\item  Introduce new randomness to keep the independence of rows in the random initialization matrices W and A at different depths. Then estimate the perturbation.
	\item Show the "off-target" Backward Correlation is zero.
	\item  Estimate the "on target" Backward Correlation by introducing a normalized input sequence $x^{(0)}$.
	\item  Explicitly construct the approximation.
\end{enumerate}

These steps  strongly rely on the normalized condition $||X_l||\ll 1 $ and this is  apparently  unrealistic. Instead, we  calculate the kernel matrix and  we introduce many new estimation to avoid this condition.

We should note that this expression $$f_L\approx\sum_lBack^{(0)} \cdot \mathbbm{1}_{\langle W, h_{l-1}\rangle+AX_l\geq 0} W^* \cdot h_{l-1}$$ is additive in itself. Thus  the nonlinear interaction between different positions  considered in this paper, especially N-variable target functions, {\bf cannot  be deduced} using the from this  method. In the previous proof,  \cite{allenzhu2019sgd}  is to use these steps to reduce the RNN function to a summation of two-layer networks  and ignore the correlation between inputs from different locations and this heavily relies on the normalized condition. In our method, we need  to consider the information in $\text{Back}$ to show the non-linear  correlation between the inputs at different positions and prove  N-variable target functions are learnable, while \cite{allenzhu2019sgd}.  requires  the normalized condition to make sure  $\text{Back}\approx \text{Back}^{(0)}$ to be roughly a constant. This is one of the most different parts between this work and  \cite{allenzhu2019sgd}.

In our case, since we do no use the normalized condition, we must show the polynomial decay of the constant part in $Back$. As mentioned in Remark 4.2, in our case, it is generally  non-trivial to show $\sqrt{\widetilde{y}^T (\bm{H}^\infty)^{-1}\widetilde{y}}\leq O(\mathscr{C}^*)$ with $\mathscr{C}^*$ polynomial in $L$. Our methods rely on a detailed estimation on the degeneracy of long RNN based on Theorem \ref{tt5}.

\section{Related Work}
{\bf Overparameterized neural network.}
In \cite{tian2017symmetry-breaking} and \cite{du2018when}, it is shown that, for a  single-hidden-node ReLU network, under mild assumptions, the loss function is one point convex in a very large area. However, in \cite{safran2018spurious}, the authors pointed out that such good properties are rare for networks with multi-hidden nodes, and indicated that an over-parameterization assumption is necessary. Similarly, \cite{2016Gradient} showed that over-parameterization can help in the training process of a linear dynamic system i.e., linear RNN. A different way to show over-parameterization is important as in  \cite{freeman2016topology}, this work proved that in the two-layer case if the number of the hidden nodes is large enough, the sub-level sets of the loss will be nearly connected. Their method can also be applied to deep networks with a skip connection in \cite{ijcai2020-387} to study the properties of loss surfaces.

Recent breakthroughs were made in understanding the neural tangent kernel(NTK)  \cite{NEURIPS20185a4be1fa, alemohammad2021the} of the neural network near the area of the random initialization. In \cite{NEURIPS201854fe976b}, \cite{DBLP}, \cite{allenzhu2019convergence} and \cite{rate}, it is shown that deep networks with a large hidden size can attain zero training error, under some assumptions of input non-degeneracy. This explains the empirical results \cite{zhang2017understanding} that DNN can fit training data with even random labels.

There are also some provable convergence results with over-parameterization going beyond NTK. The loss surface of the two-layer over-parameterized network with quadratic activation function was studied in \cite{du2018on} and \cite{MahdiTheoretical}. They showed that all the bad local minima are eliminated by over-parameterization. For ReLU activation function, in \cite{NEURIPS2019_5857d68c}, it is shown that there exits some functions can not be learned by any kernel functions but learnable with less error by a network with a skip connection. \cite{ li2020learning} provided a convergence result for learning a specific two-layer neural network which can not be learned by any kernel method, including Neural Tangent Kernel.

{\bf Generalization Ability of Deep Learning}

Classical VC theory cannot explain the generalization ability of deep learning because the VC-dimension of neural networks is at least linear in the number of parameters \cite{BartlettHLM19}. Recently, \cite{NEURIPS201962dad6e2}  showed that overparameterized neural networks can learn some notable concept classes of target functions with rich types. Moreover, their work goes beyond the NTK linearization and provides new results on  the non-convex
interactions of the three-layer network. Meanwhile, \cite{sa} provided a fine-grained analysis on the generalization error and showed the connections to the matrix of the neural tangent kernel. The results were generalized to the multi-layer case in \cite{cao}. Similar results were also studied in \cite{ji2020polylogarithmic} and \cite{chen2020overparameterization}.

Ref. \cite{NEURIPS201962dad6e2} also considered the generalization error bounds beyond the first-order NTK. It has been shown in \cite{NEURIPS201962dad6e2} that a three-layer ReLU network can provable learn some notable composite functions and dropout can help to reduce  the Rademacher  Complexity of the network thus reduce the generalization error bounds. The proof is based on the second-order NTK expansion and saddle points escaping arguments. Higher-order NTK are also studied in  \cite{Bai2020Beyond} with provable generalization error bounds. Moreover, it is shown in \cite{NEURIPS2020_fb647ca6} that comparing with the general NTK, deep networks with neural representation can achieve improved sample complexities, while for the first-order NTK, depth may not provide benefits for the learning ability \cite{bietti2021deep}.

\section{Conclusion and Future Work}
In this paper, we studied the problem of what type of function can be learned by RNN. In this work, we showed that RNNs can provably learn the two types of functions, the additive concept class and the N-variables concept class in \emph{almost-polynomial in input length many iterations and samples} starting from random initialization. For the additive concept class, we proved the result without the normalized condition and showed the almost-polynomial complexity in input length $L$. For the N-variable concept class, we showed that RNN with ReLU activation function can provably learn functions like $\psi(\langle \beta,[X_{l_1},...,X_{l_N}]\rangle )$. The complexity of learning such functions grows exponentially with either $N$ or $l_0=\max(l_1,...l_N)- \min(l_1,...l_N)$, but when one of them is small, the complexity is almost-polynomial in the input length $L$.

One of the limitations is that this work relies on the NTK linearization of RNN. One probably direction is to consider the non-convex interactions in RNN and learn more complex functions using the method in \cite{NEURIPS201962dad6e2}. Meanwhile, this work studied RNN with ReLU activation function. This did not consider the ``gate'' structure in RNN. We believe that a study on GRU, LSTM, and MGU may lead to learning more complex functions with long-term memory.

\section*{Acknowledgement}
We would like to thank Professor Wenyu Zhang for his valuable discussion, and Shuai Wang for the great help in writing. We also thank the anonymous reviewers and area chair for their helpful comments.
This research was funded by the Fundamental Research Funds for the Central Universities (Grant number 2020YJS012).

\bibliographystyle{apalike}
\bibliography{References}

\newpage
\appendix
{\bf \LARGE Supplementary Materials}

\section{Flowchart of the Proofs}

\begin{figure}[h]
\begin{tikzpicture}[->,>=stealth',shorten >=1pt,auto,node distance=1cm,
                    semithick]
  \tikzstyle{every state}=[rectangle,
                                    thick,
                                    minimum size=0.5cm,
                                    draw=black!100,
                                    fill=blue!40]
  \tikzstyle{state2}=[rectangle,
                                    thick,
                                    minimum size=0.5cm,
                                    draw=black!100,
                                    fill=green!30]
  \tikzstyle{matrix}=[rectangle,
                                    thick,
                                    minimum size=1cm,
                                    draw=black!100,
                                    fill=red!40]

  \node[state]         (t5) at (5, 5)          {\bf Theorem \ref{tt5}};
  \node[state]         (t15) at (0, 7)         { \bf Theorem \ref{t2}};

  \node[state]         (t4) at (0, 3)          {\bf Theorem \ref{mc}};
  \node[state]         (t6) at (-5, 4)        {\bf Theorem \ref{tt6}};
  \node[state]         (t17) at (-5, 6)        {\bf Theorem \ref{t17}};

  \node[state2]         (t3) at (0, -1)   {\bf Theorem \ref{mt}};
  \node[matrix]         (t2) at (0, 1)           {\bf Main Result: Theorem \ref{tt}};

  \node[state2]         (l13) at (-5, -1)   {\bf Lemma \ref{lem1}};
  \node[state2]         (l14) at (5, -1)   {\bf Lemma \ref{lem2}};
  \node[state2]         (t12) at (0, -3)   {\bf Theorem \ref{lf}};
  \node[state]         (t19) at (0, 5)   {\bf Theorem \ref{tlast}};

  \draw[->] (t15) to [out=0, in=135] node[right] {Re-Randomization}(t5);
  \draw[->] (t17) to [out=-90, in=90] node[align=center]{Key Estimation on\\ Backward Correlation}(t6);
  \draw[->] (t19) to [out=-90, in=90] (t4);
  \draw[->] (t12.north) to [out=90, in=-90] node[right] {Neural Tangent Kernel}(t3.south);
  \draw[->] (t12.west) to [out=-180, in=-45] (l13);
  \draw[->] (t12) to [out=0, in=-135] (l14);
  \draw[->] (l13) to [out=0, in=180] (t3);
  \draw[->] (l14) to [out=-180, in=0] (t3);
  \draw[->] (t5) to[out=-135,in=0] node[right]{Forward} (t4);
  \draw[->] (t6) to[out=0,in=-180] node[left]{Backward }(t4);
  \draw[->] (t4) to[out=-90,in=90] node{Complexity}(t2);
  \draw[->] (t3) to[out=90, in=-90] node[right] {Generalization Error} (t2);
  \draw[->] (t15) to[out=-90, in=90]  (t19);
  \draw[->] (t5) to[out=-180, in=0]  (t19);
  \draw[->] (t6) to[out=0, in=-180]  (t19);
  \draw[->] (t15) to[out=-180, in=30]  (t17);
\end{tikzpicture}
\caption{Flowchart of the Proof.}
\end{figure}
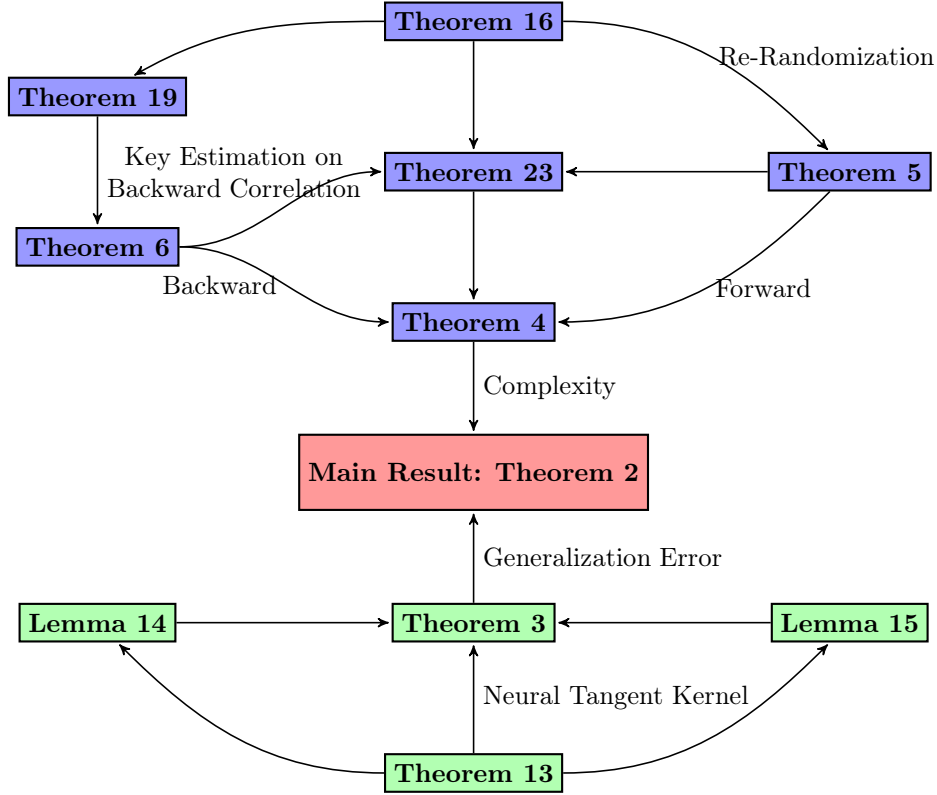
The Flowchart of the proof is shown in Figure 1. There are two parts. The first part is to prove Theorem \ref{mt}. This is easy by using techniques in \cite{rate} and \cite{cao}. The more important part is to prove Theorem \ref{mc}. We study the forward and backward correlation in Theorem \ref{tt5} and \ref{tt6}. In Theorem \ref{t17}, we show the polynomial degeneration of backward correlation which is crucial to show the complexity is polynomial in $L$.

\section {Some Probability Theory Lemmas}
\begin{def1}
A random variable $X$ is said to be sub-Gaussian with variance proxy $\sigma^2$ if $ \mathbb{E}[X]=0$ and for all $s\in\mathbb{R}$,
\begin{equation}
\mathbb{E}[e^{sX}]\leq e^{\frac{\sigma^2s^2}{\sigma^2}}.
\end{equation}
A random variable X is said to be $\lambda$-sub-exponential if $ \mathbb{E}[X]=0$, and for all $s$ that $|s|\leq \frac{1}{\lambda}$,
\begin{equation}
\mathbb{E}[e^{sX}]\leq e^{\frac{s^2\lambda^2}{2}}
\end{equation}
\end{def1}
For  $\lambda$-sub-exponential  random variable, we have the following standard concentration inequality from Chernoff bound estimation(c.f. \cite{2013Concentration}):
\begin{theo}\label{ci}
Let $X_1,X_2,...X_m$ be i.i.d $\lambda$-sub-exponential  random variable with $\lambda<\cO(1)$. Let $0<\epsilon\leq 1$. With probability at least $1-exp[\Omega(m\epsilon^2)]$,
\begin{equation}
|\frac{1}{m}\sum_{i=1}^mX_i|\leq \epsilon
\end{equation}
\end{theo}
Let $\phi$ be a function with either $|\phi(x)|\leq |Bx|$ or $|\phi(x)|\leq B$  for some $B>0$. Assuming $w$ is a Gaussian random vector, we can show $\phi(w^TX_1)\phi(w^TX_2)-\mathbb{E}\phi(w^TX_1)\phi(w^TX_2)$ is $\lambda$-sub-exponential for some $\lambda$ by estimating the moments. For $\mathbb{E}\phi(w^TX_1)\phi(w^TX_2)$, there is an equation which is a direct corollary of  Claim 4.3 in \cite{171100501}:
\begin{theo}\label{nei}
Consider $M\in \mathbb{R}^d$, all  the entries of $M$ are i.i.d. generated from $N(0,\frac{1}{d})$, and $X_1,X_2\in\mathbb{R}^d$ with $||X_1||=||X_2||=1$. Let $\mu_i(\phi)$ denote the $i-$th Hermite coefficient of function $\phi$, i.e.
$\mu_r(\phi)=\frac{1}{\sqrt{2\pi}}\int_0^\infty \phi(x)h_r(x)e^{-\frac{x^2}{2}}dx$, $h_r(x)=\frac{1}{\sqrt{r!}}(-1)^re^{\frac{x^2}{2}}\frac{d^r}{dx^r}e^{-\frac{x^2}{2}}.$

We have
\begin{equation}
\underset{M}{\mathbb{E}} \phi_1(M^TX_2)\phi_2(M^TX_1)=\sum_r \mu_r(\phi_1)\mu_r(\phi_2)(X_1^TX_2)^{r}.
\end{equation}

\begin{equation}
\underset{M}{\mathbb{E}} \phi(M^TX_2)\phi(M^TX_1)=\sum_r \mu^2_r(\phi)(X_1^TX_2)^{r}.
\end{equation}
\end{theo}

Combine the above two theorems and set $\epsilon=\frac{\log m}{\sqrt{m}}$. We have:
\begin{theo}\label{fj}
Let $\bm{W}\in \mathbb{R}^{m\times d}$. All  the entries of $M$ are i.i.d. generated from $N(0,\frac{2}{m})$, and $X_1,X_2\in\mathbb{R}^d$ with $||X_1||=||X_2||=1$. $\phi(x)=\max(0,x)$ denotes the ReLU activation function. $\mu_i(\phi)$ denotes the $i-$th Hermite coefficient of function $\phi$. $W_i$ denotes the $i$-th row of $\bm{W}$.
With probability at least $1-exp(-\Omega(\log^2m ))$,
\begin{equation}
\begin{aligned}
\phi^T(\bm{W}X_1)\phi(\bm{W}X_2)&=\sum_i \phi^T(W_iX_1)\phi(W_iX_2)\\
&=\mathbb{E}_{w\sim N(0, I_d)}\phi(w^TX_1)\phi(w^TX_2)\pm \cO(\frac{\log m}{\sqrt{m}})\\
&=\sum_r \mu_r(\phi)\mu_r(\phi)(X_1^TX_2)^{r} \pm \cO(\frac{\log m}{\sqrt{m}}).
\end{aligned}
\end{equation}
\end{theo}
This theorem is a direct corollary of the concentration inequality for the sub-exponential random variable $\phi(w^TX_1)\phi(w^TX_2)$.

In the case of ReLU function and its derivative, we can obtain analytical expressions which have been proved in \cite{NIPS2016_abea47ba,huang2020deep}:
\begin{theo}\label{fj2}
Consider functions $\phi_1(x)=\sqrt{2}\max(0,x)$ and $\phi_2(x)=\sqrt{2}\mathbbm{1}\{x>0\}$. Let $X_1, X_2\in \mathbb{R}^d, ||X_1||= ||X_2||=1$, $z=X_1^TX_2$.
\begin{equation}
\mathbb{E}_{w\sim N(0, I_d)}\phi_1(w^TX_1)\phi_1(w^TX_2)=\frac{\sqrt{1-z^2}+(\pi-arccos(z))z}{\pi},
\end{equation}
and
\begin{equation}
\mathbb{E}_{w\sim N(0, I_d)}\phi_2(w^TX_1)\phi_2(w^TX_2)=\frac{\pi-arccos(z)}{\pi}.
\end{equation}
\end{theo}
For such functions $f(z)=\mathbb{E}_{w\sim N(0, I_d)}\phi(w^TX_1)\phi(w^TX_2)$, we can see $f(0)=\mu^2_0(\phi)$ and $f'(0)=\mu^2_1(\phi)$.

\section{Technical Lemmas for RNN}
Consider equations
\begin{equation}
\begin{aligned}
&h_{l}(\bm{W},x)=\phi(\bm{W}h_{l-1}+\bm{A}X_l),\\
&f(\bm{W}),x)=\bm{B}^Th_L(x),\\
&\nabla f(\bm{W},x_i)=\sum_{l=1}^L\text{Back}^T_lD_l\cdot h_l^T(x_i),\\
&\text{Back}_l(\bm{W},x_i)=\bm{B}^TD_L\bm{W}\cdots D_{l+1}\bm{W}.
\end{aligned}
\end{equation}
The properties of $\nabla f(\bm{W},x_i)$ and $h_l$ have been already appeared in \cite{rate}. We list the results we used in this section.

Let $\bm{W}_0$ be the point of Randomly Initialization and $B(\bm{W}_0,\omega)=\{\bm{W}|\ ||\bm{W}-\bm{W}_0||_F\leq \omega\}$. We have:
\begin{lem}\label{bt}
For fixed vector $x\in \mathbb{R}^d$, $y, z\in \mathbb{R}^m$. With probability at least $1-exp(-\Omega(m/L^2))$
\begin{equation}
||\mathbbm{1}_{\bm{W}_0y+\bm{A}x>0} \cdot \bm{W}_0z||_2\leq ||z||_2(1+1/100L).
\end{equation}
For fixed $x\in \mathbb{R}^d$ and all $y,z$:
\begin{equation}
||\mathbbm{1}_{\bm{W}_0y+\bm{A}x>0} \cdot \bm{W}_0z||_2\leq ||z||_2(1+1/50L).
\end{equation}
\end{lem}
The first equation is from Claim B.13 in \cite{rate}. The second one can be easily deduced from a $\epsilon$-net argument.

\begin{lem}(Section B and Section C in \cite{rate})\label{elem}
Let $ \tau_0\leq poly(n,L), \omega\leq  \tau_0\cdot m^{-1/2}$ , $m\geq poly(L,n,\delta^{-1})$. With probability at least $1-\delta$, for all $i\in [n]$, all $l$, and $\bm{W} \in B(\bm{W}_0,\omega)$
\begin{enumerate}[(a)]
\item $||h_l(\bm{W},x_i)||\leq \cO(l)$,
\item $||\text{Back}_l(\bm{W}, x_i)D_l(\bm{W}, x_i)||_2\leq \cO(L^7\sqrt{m})$,
\item $||\bm{W}D_{l_1-1}...\bm{W}(D_{l+1})||\leq \cO(L^7)$,
\item For any vector $v$ with $||v||_0\leq\cO(L^{10/3}\tau_0^{2/3}m^{2/3})$, $||B^T(D^0_L)\bm{W}^0...\bm{W}^0v|| \leq  \sqrt{m}L^{5/3}\tau_0^{1/3}\log m\cdot m^{-1/6} $,
\item $ ||D_l'||_0\leq  \cO(L^{10/3}\tau_0^{2/3}m^{2/3}).$
\end{enumerate}
\end{lem}
The (a) is from the proof of Lemma B.3 and Lemma C.2a in \cite{rate}, and the (b) is from Lemma C.9 and Lemma B.11  in \cite{rate}. (c) is from Lemma C.7 in \cite{rate}. From Corollary B.18, Lemma C.11 and Claim G.2 in \cite{rate} we have (d) and (e).

In our case, $||\bm{A}X_l||\leq \frac{1}{L^{3/2}}$, rather than  $||\bm{A}X_l||\leq \cO(1)$. These bounds can be improved, but since we mainly care about the exponential dependence on $L$, we do not use it.

These equations deduce the following linearization theorem which is an analogue of Lemma 4.1 in \cite{cao}:
\begin{theo}\label{lf}
With probability at least $1-\cO(n)\cdot exp(-\Omega(\log m))$, for all $i\in [n]$ and $\bm{W},\bm{W}'\in B(\bm{W}_0,\omega)$,
\begin{equation}
|f(\bm{W}',x_i)-f(\bm{W},x_i)-\langle \nabla f(\bm{W},x_i), \bm{W}'-\bm{W}\rangle |\leq \cO(\omega^{1/3}L^{10}\log m \sqrt{m})||\bm{W}'-\bm{W}||_2.
\end{equation}
\end{theo}
{\bf Proof:}
Let
\begin{equation}
\begin{aligned}
&h_L(x)=h_L(\bm{W},x),\  h_L(\bm{W'},x)=h_L(x)+h'_L(x),\\
&D_l=D_l(\bm{W},x),\ D_l'=D_l(\bm{W}',x), \ D^0_l=D_l(\bm{W}_0,x).
\end{aligned}
\end{equation}
By Claim G.2 in \cite{rate}, there exits diagonal matrices $D''_l$, $\{D''_l\}_{ii}\neq 0$ if and only if $ \{D'_l\}_{ii}\neq 0$, $||D''_l||_0\leq ||D_l'||_0\leq \cO(L^{10/3}\tau_0^{2/3}m^{2/3})$, and
\begin{equation}
\begin{aligned}
B^T(h_L(x)+h'_L(x))-B^Th_L(x)=&\sum_{l=1}^{L-1}B^T(D_L+D''_L)\bm{W}'...(D_{l+1}+D''_{l+1})\\
&\cdot (\bm{W}'-\bm{W})h_l(x).
\end{aligned}
\end{equation}

Then,
\begin{equation}
\begin{aligned}
&f(\bm{W}',x_i)-f(\bm{W},x_i)-\langle \nabla f(\bm{W},x_i), \bm{W}'-\bm{W}\rangle\\
&=\sum_{l=1}^{L-1}B^T(D_L+D''_L)\bm{W}'...(D_{l+1}+D''_{l+1})\cdot (\bm{W}'-\bm{W})h_l(x)\\
&-B^TD_L\bm{W}...\bm{W}D_{l+1}\cdot (\bm{W}'-\bm{W})h_l(x).
\end{aligned}
\end{equation}
To prove the theorem, same as Lemma 5.7 in \cite{allenzhu2019convergence}, we have the following result:
Let $D^{0/1}_l$  be diagonal matrix and $(D^{0/1}_l)_{ii}=0$ if $(D_l+D''_l-D^0_l)_{ii}=0$,  $(D^{0/1}_l)_{ii}=1$ if $(D_l+D''_l-D^0_l)_{ii}\neq 0$. With probability at least $1-\delta$,
\begin{equation}
\begin{aligned}
&||B^T(D_L+D''_L)\bm{W}'...(D_{l+1}+D''_{l+1})\cdot \bm{W}')-B^TD_L\bm{W}...\bm{W}D_{l+1}\cdot \bm{W}||\\
&\leq \cO(\sum_{l_1=l+1}^L\underbrace{||B^T(D^0_L)\bm{W}^0...\bm{W}^0(D^{0/1}_{l_1})||}_{T_1}\cdot ||D''_{l_1}||\cdot \underbrace{||(D^{0/1}_{l_1})\bm{W}'D'_{l_1-1}...\bm{W}'(D'_{l+1})||}_{T_2})\\
&\overset{(a)}{\leq} \cO(\sqrt{m}L^{5/3+8}\tau_0^{1/3}\log m\cdot m^{-1/6})\leq \cO(\sqrt{m}L^{10}\omega^{1/3}\log m).
\end{aligned}
\end{equation}
In (a), $T_2\leq O(L^7)$ is from (c) in Lemma \ref{elem}.  From (d) in Lemma \ref{elem} and $||D''_l||_0\leq   \cO(L^{10/3}\tau_0^{2/3}m^{2/3})$, $T_1\leq \sqrt{m}L^{5/3}\tau_0^{1/3}\log m\cdot m^{-1/6}$.
\QEDA

\begin{rem}
In this theorem,
\begin{equation}
|f(\bm{W}',x_i)-f(\bm{W},x_i)-\langle \nabla f(\bm{W},x_i), \bm{W}'-\bm{W}\rangle |\leq \cO(\omega^{1/3}L^{10}\log m \sqrt{m})||\bm{W}'-\bm{W}||_2.
\end{equation}
And in \cite{cao}, there is a similar result that
\begin{equation}
|f(\bm{W}',x_i)-f(\bm{W},x_i)-\langle \nabla f(\bm{W},x_i), \bm{W}'-\bm{W}\rangle |\leq \cO(\omega^{1/3}L^2\sqrt{\log m} \sqrt{m})||\bm{W}'-\bm{W}||_2.
\end{equation}
The differences on $\log m$ are from that Lemma 4.4 in \cite{allenzhu2019convergence} says if $||u||_0\leq s$, $|B^T(D_L+D''_L)\bm{W}'_L...(D_{l+1}+D''_{l+1})\cdot \bm{W}'_{l+1}u|\leq\cO(\sqrt{s \log m}$
and Corollary B.18 in \cite{rate} says $|B^T(D_L+D''_L)\bm{W}'...(D_{l+1}+D''_{l+1})\cdot \bm{W}'u|\leq\cO(\sqrt{s}\log m)$ for RNN case.
\end{rem}

\section{Generalization properties: Proof of Theorem \ref{mt}}
\begin{lem}\label{lem1}
Denote $L_i(\bm{W})=\ell(y_i\cdot f(\bm{W},x_i))$. Suppose there exits $\bm{W}^* \in B(\bm{W}_0,R/\sqrt{m})$ with $R\leq poly(n,L)$, $L_i(\bm{W}^*)\leq \frac{1+R^2}{n}$. For any $ \delta$, there exists
\begin{equation}
m^*(n,\delta,R,L)=poly(n,R,L,\delta^{-1})
\end{equation}
such that if $m>m^*$, with probability at least $1-\delta$,  SGD with $\eta=1/m$ for some small enough $\nu$ will output:
\begin{equation}
\frac{1}{n}\sum_{i=1}^n L^{0-1}_{\emph{D}}(\bm{W}^{i})\leq \cO(\frac{1}{ n})+ \cO(\frac{R^2}{ n})+\cO(\frac{\log(1/\delta)}{n}).
\end{equation}
\end{lem}

{\bf Proof of Lemma \ref{lem1}:}

Firstly, for all $i$, $\bm{W}\in B(\bm{W}_0, \omega), \omega\leq R/m^{1/2}$, from Lemma \ref{elem}, $||\nabla f(\bm{W},x_i)||_F\leq \cO(L^{8}\sqrt{m})$.
\begin{equation}
||\bm{W}^{i+1}-\bm{W}^0||_F\leq \sum_{k=1}^i||\bm{W}^{k+1}-\bm{W}^k||_F\leq \cO(n\eta L^{8}\sqrt{m})\leq \frac{L^{8} R}{\sqrt{m}}\leq \cO(\tau_0/m^{1/2})
\end{equation}
with $\tau_0\leq poly(n,L)$.
Thus we can use Theorem \ref{lf}. We have,
\begin{equation}
\begin{aligned}
L_i(\bm{W}^{i})-L_i(\bm{W}^*)&\leq \langle \nabla_{\bm{W}}L_i(\bm{W}^{i}), \bm{W}^{i}-\bm{W}^*\rangle \\
&+ |\ell'(y_if(\bm{W},x_i))\cdot y_i|\cdot \cO(\omega^{1/3}L^{10}\log m\sqrt{m})||\bm{W}^i-\bm{W}^*||_2\\
&=\frac{\langle \bm{W}^i-\bm{W}^{i+1}, \bm{W}^i-\bm{W}^*\rangle}{\eta}  + \cO(\omega^{1/3}L^{10}\log m\sqrt{m})||\bm{W}^i-\bm{W}^*||_2
\end{aligned}
\end{equation}
Therefore,
\begin{equation}
\begin{aligned}
\sum_{i=1}^n L_i(\bm{W}^{i})&\leq \sum_{i=1}^n \{L_i(\bm{W}^*) +\frac{R^2}{2\eta m} + \cO(\omega^{1/3}L^{10}\log m\sqrt{m})\sum_{i=1}^n||\bm{W}^i-\bm{W}^*||_2\},\\
&\leq \sum_{i=1}^n\{L_i(\bm{W}^*) +\frac{R^2}{2\eta m} + \cO(L^{10}\log m\cdot n\cdot R^{4/3}\cdot m^{-1/6})\},\\
&\overset{(a)}{\leq} \sum_{i=1}^nL_i(\bm{W}^*) +  R^2.\\
\end{aligned}
\end{equation}
In (a), we use $m>m^*\sim poly(n,L)$.

Therefore,
\begin{equation}
\begin{aligned}
&\frac{1}{n}\sum_{i=1}^n L_i(\bm{W}^{i})\leq  \frac{1+R^2}{n} +  \frac{R^2}{n}.
\end{aligned}
\end{equation}
The cross-entropy function $\ell(x)$ satisfies that $L'_i(\bm{W}^{i}) \leq L_i(\bm{W}^{i})$ and $L^{0-1}_i(\bm{W}^{i})\leq L'_i(\bm{W}^{i}) $, where
\begin{equation}
L'_i(\bm{W}^{i})=-\ell'(y_if(\bm{W},x_i)).
\end{equation}
And $-\ell'(x)$ is bounded. Using the boundedness and a martingale Bernstein bound argument as Lemma 4.3 in \cite{ji2020polylogarithmic}, we have
\begin{equation}
\begin{aligned}
&\frac{1}{n}\sum_{i=1}^n L^{0-1}_D(\bm{W}^{i})\leq   \cO(\frac{1}{ n})+\cO(\frac{R^2}{n})+\cO(\frac{\log(1/\delta)}{n}).
\end{aligned}
\end{equation}
\QEDA
\begin{rem}
The result of generalization error $1/n$ is this better than that in \cite{cao} $1/\sqrt{n}$, which shows
\begin{equation}
\begin{aligned}
&\frac{1}{n}\sum_{i=1}^n L^{0-1}_D(\bm{W}^{i})\leq  \frac{4}{n}\sum_{i=1}^nL_i(\bm{W}^*) +  \cO(\frac{R}{\sqrt{n}})+\cO(\sqrt{\frac{\log(1/\delta)}{n}}).
\end{aligned}
\end{equation}
This is because Lemma 4.3 in \cite{ji2020polylogarithmic}  makes use of the boundedness of $L_i(\bm{W})$. Thus it is  applicable in this theorem. There is also a similar argument in Lemma 5.6 of  \cite{chen2020overparameterization}.
\end{rem}

\begin{lem}\label{lem2}
Under the condition of Theorem \ref{mt}, with probability at least $1-\delta$, there exits $\bm{W}^* \in B(\bm{W}_0,R/\sqrt{m})$, such that $L_i(\bm{W}^*)\leq \frac{1+R^2}{n}$, $R\leq\widetilde{\cO}(L\sqrt{\widetilde{y}^T(\bm{H}^{\infty})^{-1}\widetilde{y}})$.
\end{lem}

{\bf Proof of Lemma \ref{lem2}:}\\
Let $\bm{\epsilon}$ be the matrix in (\ref{condition1}),
$$\bm{G}=m^{-1/2}\cdot (vec[\nabla f(\bm{W}^0,x_1)], vec[\nabla f(\bm{W}^0,x_2)],... vec[\nabla f(\bm{W}^0,x_n)])\in \mathbb{R}^{m^2\times n}.$$
 \begin{equation}\label{svd}
\bm{G}+\bm{\epsilon}=\bm{P}\bm{\Lambda}\bm{Q}^T.
\end{equation}
 is the singular value decomposition. Note that $m^2\gg n$. We can set $ \bm{\epsilon}^T \bm{G} =0$ without changing $\bm{\epsilon^T\epsilon}$.

With probability at least $1-\delta$, for all $i\in [n]$, $ |f(\bm{W}^0,x_i)|\leq \cO(L\log(n/\delta))$.
 We assume $w^*=\bm{P}\bm{\Lambda}^{-1}\bm{Q}^T(B\cdot \widetilde{y})$, with $0<|f(\bm{W}^0,x_i)|+\log\{1/[exp(n^{-1})-1]\}+0.01<B\leq \cO(L\log(n/\delta))$ for all $i\in [n]$, then $||w^*||^2_2\leq B^2\widetilde{y}^T ( \bm{H}+\bm{\epsilon}^T\bm{\epsilon})^{-1} \widetilde{y}.$ and $\bm{G}^Tw^*=B\cdot \widetilde{y}-\bm{\epsilon}^Tw^*$. Meanwhile, reshape $w^*$ as $\bm{W}^*\in \mathbb{R}^{m\times m}$, then we have
 \begin{equation}
 \langle \nabla f(\bm{W}^0,x_i), \bm{W}^*-\bm{W}^0 \rangle =  B\cdot \widetilde{y}_i \pm||\bm{\epsilon}||_F \cdot \sqrt{\widetilde{y}^T (\bm{H}+\bm{\epsilon})^{-1}\widetilde{y}} = B\cdot \widetilde{y}_i \pm 0.01.
 \end{equation}
 Therefore $\bm{W}^* \in B(\bm{W}_0,\widetilde{O}(L\mathscr{C}^*/\sqrt{m}))$, and
 \begin{equation}
 \begin{aligned}
 \ell(y_i\cdot(f(\bm{W}^*,x_i)))&\leq\ell(y_i\cdot \{f(\bm{W}^0,x_i)+\langle \nabla f(\bm{W}^0,x_i), \bm{W}^*-\bm{W}^0 \rangle\}) \\
 &+|\ell'(y_if(\bm{W},x_i))\cdot y_i|\cdot \cO(L^{10}\log m\cdot n\cdot R^{4/3}\cdot m^{-1/6})\\
 &\leq\ell(y_i\cdot \{f(\bm{W}^0,x_i)+\langle \nabla f(\bm{W}^0,x_i), \bm{W}^*-\bm{W}^0 \rangle\}) \\
 &+R^2/n\\
 &\leq \ell(\log(1/[exp(n^{-1})-1])) +R^2/n,\\
 & \leq n^{-1}+R^2/n.
 \end{aligned}
 \end{equation}
Thus $L_i(\bm{W}^*)\leq \frac{1+R^2}{n}$.
\QEDA

Then Theorem \ref{mt} follows from Lemma \ref{lem1} and \ref{lem2}.

\section {Forward Correlation: Proof of Theorem \ref{tt5}}
\begin{theo}\label{t2}
Consider equation $h'_l(x_1)=\phi(\bm{W}_lh'_{l-1}(x_l)+\bm{A}_lX_l)$, where the entries of $\bm{W}^l$ and $\bm{A}^l$ are i.i.d. generated from $N(0,\frac{2}{m})$ and $N(0,\frac{2}{L^3m})$.  $\bm{W}^l$, $\bm{A}^l$ and $\bm{W}^{l'}$, $\bm{A}^{l'}$ are independent when $l\neq l'$. With probability at lesat $1-L^2exp(-\Omega(\log^2m))$. For all $1<l\leq L$, we have
\begin{equation}
|h_{l}^T(x)h_{l}(x')-  {h'_{l}}^{T}(x)h'_{l}(x')|\leq \cO(L^{2}\log^2m/\sqrt{m})
\end{equation}
for  $x,x'=x_1,x_2$.
\end{theo}

In order to prove the theorem, firstly we claim that
\begin{lem}\label{l3}
Let $h_l(x)=\phi(\bm{W}h_{l-1}(x)+\bm{A}X_l)$. $\widetilde{h}_l(x)=\phi(\widetilde{\bm{W}}\widetilde{h}_{l-1}(x)+\widetilde{\bm{A}}X_l)$ is defined by  $\widetilde{\bm{W}}, \widetilde{\bm{A}}$. $\widetilde{\bm{W}}, \widetilde{\bm{A}}$ and $\bm{W}, \bm{A}$ are i.i.d. Then for any $0<l,l'<L$, with probability at least $1-L^2exp(-\Omega(\log^2m))$,
\begin{equation}
|h_{l}^T(x)h_{l'}(x')-  {\overline{h}_{l}}^{T}(x)\overline{h}_{l'}(x')|\leq \cO(l^{2}\log^2m/\sqrt{m})
\end{equation}
where
$$\overline{h}_l(x)=\phi(\bm{W}\widetilde{h}_{l-1}(x)+\bm{A}X_l)$$
$$\overline{h}_{l'}(x')=\phi(\bm{W}\widetilde{h}_{l'-1}(x')+\bm{A}X'_l)$$
\end{lem}

{\bf Proof of Theorem \ref{t2}:}\\
In the case $l=1$, $h_{1}(x)=\phi(\bm{W}h_0+\bm{A}X_1)$.

From Theorem \ref{fj} we have, with probability at least $1-exp(-\Omega(\log^2m))$
\begin{equation}
h_{1}^T(x)h_{1}(x')=\mathbb{E}h_{1}^T(x)h_{1}(x')\pm \cO(\log^2m/m)={h'_{1}}^T(x)h'_{1}(x')\pm \cO(\log^2m/\sqrt{m})
\end{equation}

The theorem is true.

Supposing the theorem is true for $l$, for $l+1$, using Lemma \ref{l3}

\begin{equation}
 h_{l+1}^T(x)h_{1+1}(x')=\mathbb{E}\phi(\bm{W}\widetilde{h}_{l}(x')+\bm{A}X'_{l+1})\phi(\bm{W}\widetilde{h}_{l}(x)+\bm{A}X_{l+1})\pm \cO(l^2\log^2m/\sqrt{m})
\end{equation}

\begin{equation}
\bm{W}\widetilde{h}_{l}(x)+\bm{A}X_{l+1}=
\begin{bmatrix}
\bm{W} &L^{3/2}\bm{A}
\end{bmatrix}
\cdot
\begin{bmatrix}
\widetilde{h}_{l}(x)\\
\frac{1}{L^{3/2}}X_{l+1}
\end{bmatrix}
=\bm{M}\cdot z
\end{equation}
Thus
\begin{equation}
\begin{aligned}
&\mathbb{E}\phi(\bm{W}\widetilde{h}_{l}(x')+\bm{A}X'_{l+1})\phi(\bm{W}\widetilde{h}_{l}(x)+\bm{A}X_{l+1})\\
&=\mathbb{E}_{w\sim N(0,\sqrt{2}\bm{I}_{m+d})}\phi(\bm{w}^Tz)\phi(\bm{w}^Tz')\\
&= {h'_{l+1}}^T(x)h'_{l+1}(x')\pm \cO(l^{16}\log^2m/\sqrt{m})
\end{aligned}
\end{equation}
$$h_{l+1}^T(x)h_{1+1}(x')= {h'_{1+1}}^T(x)h'_{1+1}(x')\pm \cO((l+1)^{16}\log^2m/\sqrt{m}).$$
\QEDA

{\bf Proof of Lemma \ref{l3}:}\\
In the case $l=1$, this is true from Theorem \ref{fj}.

For $l>1$, We use the Gram-Schmidt orthonormal matrix as  Claim B.4 in \cite{rate}. let $\bm{U}_l\in \mathbb{R}^{m\times 2l}$ denote column orthonormal matrix using Gram-Schmidt as
\begin{equation}
\begin{aligned}
&\bm{U}_l=GS(h_1(x_1),h_1(x_2),...h_l(x_1),h_l(x_2)).
\end{aligned}
\end{equation}
We can write
\begin{equation}
\begin{aligned}
&Wh_{l}=WU_{l-1}U_{l-1}^Th_{l-1}+W(I-U_{l-1}U_{l-1}^T)h_l,\\
\end{aligned}
\end{equation}
and
\begin{equation}
\begin{aligned}
h_{l+1}(x)&=\phi(WU_{l-1}U_{l-1}^Th_l+W(I-U_{l-1}U_{l-1}^T)h_l+AX_{l+1})
\end{aligned}
\end{equation}

Consider
\begin{equation}
\begin{aligned}
h_{l+1}(x_2)^Th_{l+1}(x_1)=&\phi(WU_{l-1}U_{l-1}^Th_l(x_2)+W(I-U_{l-1}U_{l-2}^T)h_l(x_2)+Ax_{2,l+1})^T\\
&\cdot\phi(WU_{l-1}U_{l-1}^Th_l(x_1)+W(I-U_{l-1}U_{l-1}^T)h_l(x_1)+Ax_{1,l+1})
\end{aligned}
\end{equation}
We write $z_1=(I-U_{l-1}U_{l-1}^T)h_l(x_1)$, $z_2=(I-U_{l-1}U_{l-1}^T)h_l(x_2).$ $q_1=U_{l-1}^Th_l(x_1), q_2=U_{l-1}^Th_l(x_2)$.

\begin{equation}
\begin{aligned}
z_2=\frac{\langle z_1,z_2\rangle z_1}{||z_1||^2}+(I-z_1z_1^T/||z_1||^2)z_2
\end{aligned}
\end{equation}

Then
\begin{equation}
\begin{aligned}
h_{l+1}(x_2)^Th_{l+1}(x_1)=&\phi(WU_{l-1}q_2+W\frac{\langle z_1,z_2\rangle z_1}{||z_1||^2}+W(I-z_1z_1^T/||z_1||^2)z_2+Ax_{2,l+1})^T\\
&\cdot\phi(WU_{l-1}q_1+Wz_1+Ax_{1,l+1})
\end{aligned}
\end{equation}

Thus
\begin{equation}
\begin{aligned}
h_{l+1}(x_2)^Th_{l+1}(x_1)=&\phi(
\begin{bmatrix}
M_1 &M_2 & M_3&M_4
\end{bmatrix}
\cdot
\begin{bmatrix}
q_2\\
\frac{\langle z_1,z_2\rangle }{||z_1||}\\
||(I-z_1z_1^T/||z_1||^2)z_2||\\
x_{2,l+1}
\end{bmatrix})^T\\
&\cdot\phi(
\begin{bmatrix}
M_1 &M_2 & M_3&M_4
\end{bmatrix}
\cdot
\begin{bmatrix}
q_1\\
||z_1||\\
0\\
x_{1,l+1}
\end{bmatrix})
\end{aligned}
\end{equation}
where 
\begin{equation}
\begin{bmatrix}
M_1 &M_2 & M_3&M_4
\end{bmatrix} 
=
\begin{bmatrix}
WU_{l-1} &W z_1/||z_1|| & W(I-z_1z_1^T/||z_1||^2)z_2/||(I-z_1z_1^T/||z_1||^2)z_2||&A
\end{bmatrix} 
\end{equation}
Let 
\begin{equation}
\begin{aligned}
&E_2=\begin{bmatrix}
q_2\\
\frac{\langle z_1,z_2\rangle }{||z_1||}\\
||(I-z_1z_1^T/||z_1||^2)z_2||\\
x_{2,l+1}
\end{bmatrix})\\
&E_1=\begin{bmatrix}
q_1\\
||z_1||\\
0\\
x_{1,l+1}
\end{bmatrix})
\end{aligned}
\end{equation}
We have $E_2^TE_1=h_{l}(x_2)^Th_{l}(x_1)+x_{2,l+1}^Tx_{1,l+1}$.

Using a similar proof as  Claim B.4 and Claim B.4 in \cite{rate}, we have for any $E_1, E_2$, with probability at least $1-exp(-\Omega(\log^2m))$,
\begin{equation}
\begin{aligned}
&\phi(
\begin{bmatrix}
M_1 &M_2 & M_3&M_4
\end{bmatrix}
\cdot
E_2)^T
\cdot\phi(
\begin{bmatrix}
M_1 &M_2 & M_3&M_4
\end{bmatrix}
\cdot
E_1)\\
&=\mathbb{E}_{M\sim \mathcal{N}(0, I)} \phi(ME_2)\phi(ME_1)\pm \cO({l+1})^2\frac{\log^2m}{\sqrt{m}})
\end{aligned}
\end{equation}

Thus
\begin{equation}
\begin{aligned}
&h_{l'+1}(x')^Th_{l+1}(x)\\
=&\overline{h}_{l'+1}(x')^T\overline{h}_{l+1}(x)\pm \cO((l+1)^{2}\log^2m/\sqrt{m}).
\end{aligned}
\end{equation}

The theorem follows.
\QEDA

Combing above theorems, Theorem \ref{nei},\ref{fj} and \ref{fj2}, we have
\begin{lem}\label{io}
Let $$Q_l=\sqrt{(1+\frac{1}{L^3}\sum_{k=1}^l||X_{i,k}||^2)\cdot(1+\frac{1}{L^3}\sum_{k=1}^l||X_{j,k}||^2) }$$
 $$\Gamma(z)=\frac{\sqrt{1-z^2}+(\pi-arccos(z))z}{\pi}$$

There exits $K^l_{i,j}$ such that with probability at least $1-L^2exp(-\Omega(\log^2m))$,
\begin{equation}
|h_{l}^T(x_i)h_{l}(x_j)-K^l_{i,j}|\leq \cO( \frac{l^{2}\log^2m}{\sqrt{m}}).
\end{equation}
And
\begin{equation}
\begin{aligned}
&K^1_{i,j}= Q_1\cdot\Gamma([1+\frac{1}{L^3}X_{i,1}^TX_{j,1}]/Q_1)\\
&K^{l}_{i,j}=Q_l\cdot\Gamma(\{\frac{1}{L^3}X_{i,l}^TX_{j,l}+ K^{l-1}_{i,j}\}/Q_l)^r
\end{aligned}
\end{equation}
\end{lem}
Thus Theorem \ref{tt5} follows.
\section{Backward Correlation: Proof of Theorem \ref{tt6}}
\begin{theo}\label{t17}
For $l\neq l'$, with probability at least $1-L^2exp(-\Omega(\log^2m))$,
\begin{equation}\label{ea}
|\frac{1}{m}\langle \text{Back}_l(x_i)\cdot D_l,\text{Back}_{l'}(x_j)\cdot D_{l'}' \rangle| \leq \cO(\frac{L^4\log^4 m}{m^{1/4}}).
\end{equation}
For $l=l'$, with probability at least $1-L^2exp(-\Omega(\log^2m))$,
\begin{equation}\label{eb}
\frac{1}{m}\langle \text{Back}_l(x_i)\cdot D_l,\text{Back}_{l}(x_j)\cdot D_{l}' \rangle \succeq \Omega(1/L^4)\cdot \Sigma(\{ \frac{1}{L^3}\langle X_{i,l}, X_{j,l} \rangle +K^{l-1}_{i,j}\}/Q_l)\pm \cO(\frac{L^4\log^4 m}{m^{1/4}}).
\end{equation}
\end{theo}

{\bf Proof of (\ref{ea}): }

The proof of (\ref{ea}) is almost a line-by-line copy of the proof in section C of \cite{allenzhu2019sgd}, but there are some minor differences.

Let $\zeta_1,...,\zeta_m$  be a random orthonormal basis of $\mathbb{R}^m$. Then divide all the $m$ coordinates into $\sqrt{m}$ chunks $N_1, N_2,...N_{m^{1/2}}$ of the size $N=\sqrt{m}$.

Define
\begin{equation}
z_{1,0}=D_l\zeta_1, z'_{1,0}=D'_l\zeta_1,\ ...\ z_{N,0}=D_l\zeta_N, z'_{N,0}=D'_l\zeta_N
\end{equation}
and
\begin{equation}
\begin{aligned}
&z_{i,a}=D_{l+a}W\cdots D_{l+1}WD_lz_{i,1}\\
&z'_{i,a}=D'_{l'+a}W\cdots D'_{l+1}WD'_{l'}z'_{i,1}
\end{aligned}
\end{equation}

\begin{equation}
Z_{p,a}=GS(h_1,..., h_{\max(l,l')},z_{1,1},...,z_{N,1},z'_{1,1},...,z'_{N,1},...,z_{1,a},...,z_{p,a},z'_{1,a},...,z'_{p,a})
\end{equation}

We claim that, with probability at least $1-L^2exp(-\Omega(\log^2m))$, for all $a$,
\begin{equation}
||Z_{p,a}^Tz_{p,a}||\leq \cO(\frac{L^3\sqrt{N}\log^3m}{\sqrt{m}}).
\end{equation}
When $a=0$,
\begin{equation}
Z_{p,0}^Tz_{p,0}=Z_{p,0}^TD_l\zeta_1
\end{equation}
With probability at least $1-exp(-\Omega(\log^2m))$,
\begin{equation}
||Z_{p,0}^Tz_{p,0}||\leq \cO(l\log m/\sqrt{m}).
\end{equation}
 For $a>1$,
\begin{equation}
Z_{p,a+1}^Tz_{p,a+1}=Z_{p,a+1}^TD_{l+a+1}(W(I-Z_{p,a+1}Z_{p,a+1}^T)z_{p,a}+WZ_{p,a+1}Z_{p,a+1}^Tz_{p,a}),
\end{equation}
\begin{equation}
\begin{aligned}
&||Z_{p,a+1}^TD_{l+a+1}WZ_{p,a+1}Z_{p,a+1}^Tz_{p,a})||\leq ||D_{l+a+1}WZ_{p,a+1}Z_{p,a+1}^Tz_{p,a})||\\
&\leq ||Z_{p,a+1}^Tz_{p,a})||(1+\frac{1}{50L}),
\end{aligned}
\end{equation}
The last step is from Lemma \ref{bt}.

And
\begin{equation}
||Z_{p,a+1}^TD_{l+a+1}W(I-Z_{p,a+1}Z_{p,a+1}^T)z_{p,a}||\leq \cO(\frac{(l+a)^3\sqrt{N}\log^2m}{\sqrt{m}}).
\end{equation}
is because $W(I-Z_{p,a+1}Z_{p,a+1}^T)z_{p,a}\sim N(0,(2\bm{I}/m) \cdot ||(I-Z_{p,a+1}Z_{p,a+1}^T)z_{p,a}||^2)$.

This claim follows that,
\begin{equation}
\sum_{p\in[N]} \Xi_p=\sum_p B^T(I-Z_{p,a}Z_{p,a}^T)z_{p,a}\cdot B^T(I-Z_{p,a'}Z_{p,a'}^T)z'_{p,a'}\pm \cO(m^{1/4}L^3\log^4{m})
\end{equation}
In the case $a\neq a'$, $(I-Z_{p,a}Z_{p,a}^T)z_{p,a}$ and $(I-Z_{p,a'}Z_{p,a'}^T)z'_{p,a'}$ are mutually orthogonal. With probability at least $1-exp(-\Omega(\log^2m))$,
\begin{equation}
|\sum_p B^T(I-Z_{p,a}Z_{p,a}^T)z_{p,a}\cdot B^T(I-Z_{p,a'}Z_{p,a'}^T)z'_{p,a'}|\leq  \cO(log^4m)
\end{equation}
Thus
\begin{equation}
|\frac{1}{m}\langle \text{Back}_l(x_i)\cdot D_l,\text{Back}_{l'}(x_j)\cdot D_{l'}' \rangle| \leq \cO(\frac{L^3\log^4 m}{m^{1/4}}).
\end{equation}
There are $\sqrt{m}$ chunks, thus with probability at least $1-\sqrt{m}L^2exp(-\Omega(\log^2m))=1-L^2exp(-\Omega(\log^2m))$.
(\ref{ea}) follows.
\QEDA

{\bf Proof of (\ref{eb}):}

For any $a$, we have,
\begin{equation}
z_{p,a+1}=D_{l+a+1}(W(I-Z_{p,a+1}Z_{p,a+1}^T)z_{p,a}+WZ_{p,a+1}Z_{p,a+1}^Tz_{p,a})
\end{equation}
Thus,
\begin{equation}
||z_{p,a+1}-D_{l+a+1}W(I-Z_{p,a+1}Z_{p,a+1}^T)z_{p,a}||\leq \cO(\frac{L^3\sqrt{N}\log^3m}{\sqrt{m}})
\end{equation}

We know that
$\frac{1}{m}\langle \text{Back}_l(x_i)\cdot D_l,\text{Back}_{l}(x_j)\cdot D_{l}' \rangle=\sum_{i=1}^{m^{1/2}} \Theta_i$, where
\begin{equation}
\begin{aligned}
\Theta_i=\sum_{p\in[N_i]} \Xi_p=&\sum_p B^T(I-Z_{p,a}Z_{p,a}^T)D_{l+a}W(I-Z_{p,a}Z_{p,a}^T)...D_{l+1}W(I-Z_{p,1}Z_{p,1}^T)z_{p,0}\\
&\cdot B^T(I-Z_{p,a}Z_{p,a}^T)D'_{l+a}W(I-Z_{p,a}Z_{p,a}^T)...D'_{l+1}W(I-Z_{p,1}Z_{p,1}^T)z'_{p,0}\\
&\pm \cO(m^{1/4}L^3\log^4{m})
\end{aligned}
\end{equation}

Combine the facts :
\begin{itemize}
\item With probability at least $1-exp(-\Omega(\log^2m))$,
\begin{equation}
\sum_pB^Tz_{p,a}\cdot B^Tz'_{p,a}=\langle z_{p,a}, z'_{p,a}\rangle \pm \cO(\frac{\sqrt{N}L^2\log^2m}{\sqrt{m}}).
\end{equation}
\item Let
\begin{equation}
\begin{aligned}
&D_l=\phi(\bm{W}h_{l-1}(x_l)+\bm{A}X_l)\\
&\widetilde{D_l}=\phi(\bm{W}\widetilde{h}_{l-1}(x_l)+\bm{A}X_l)
\end{aligned}
\end{equation}
where $\widetilde{h}$ is $h_l$ define by re-randomization in Lemma \ref{l3}. Then
$|\langle D'_l, D_l\rangle- \langle \widetilde{D'_l}, \widetilde{D_l}\rangle| \leq \cO(\frac{L^2\log^2m}{m})$

\item
\begin{equation}
||Z_{p,a}^Tz_{p,a}||\leq \cO(\frac{L^3\sqrt{N}\log^3m}{\sqrt{m}}).
\end{equation}
\end{itemize}
and Claim \ref{c51}. Let $Q_l=\sqrt{(1+\frac{1}{L^3}\sum_{k=1}^l||X_{k}||^2)\cdot(1+\frac{1}{L^3}\sum_{k=1}^l||X'_{k}||^2) }$.
With probability at least $1-exp(-\Omega(\log^2m))$, we have
\begin{equation}
\begin{aligned}
\langle z_{p,a}, z'_{p,a}\rangle=&\langle (I-Z_{p,a-1}Z_{p,a-1}^T)z_{p,a-1}, (I-Z_{p,a-1}Z_{p,a-1}^T)z'_{p,a-1}\rangle\\
 &\cdot \Sigma(\{h_{l+a-1}^Th'_{l+a-1}+ \frac{1}{L^3}X^T_{l+a}X_{l+a}\}/Q_{l+a}) \pm \cO(\frac{L^3\sqrt{N}\log^3m}{\sqrt{m}}),\\
 =&\langle (z_{p,a-1}, z'_{p,a-1}\rangle\cdot \Sigma(\langle h_{l+a-1}, h'_{l+a-1} \rangle/Q_{l+a}\\
 &+  \frac{1}{L^3}\langle X_{l+a},X'_{l+a} \rangle/Q_{l+a})
  \pm \cO(\frac{L^3\sqrt{N}\log^3m}{\sqrt{m}}),
\end{aligned}
\end{equation}
where
\begin{equation}
\Sigma(x)=\frac{1}{2}+\frac{ arcsin(x)}{\pi}=\frac{\pi-arccos(x)}{\pi}.
\end{equation}
In order to study the constant term in $$\Sigma(\langle h_{l+a-1}, h'_{l+a-1} \rangle/Q_{l+a}+ \frac{1}{L^3}\langle X_{l+a},X'_{l+a} \rangle/Q_{l+a}),$$ we need to study$ \langle h_{l+a-1}, h'_{l+a-1} \rangle.$

The constant term in $\langle h_{l+a-1}, h'_{l+a-1} \rangle/Q_{l+a}$  is the sequence (Lemma \ref{io}):
\begin{equation}
\begin{aligned}
&K_l=\Gamma(K_{l-1}\cdot Q_{l-1}/Q_{l}),\\
&\Gamma(x)=x+\frac{\sqrt{1-x^2}-arccos(x)x}{\pi}.
\end{aligned}
\end{equation}
Note that $K_l>0$ is convergent. Meanwhile, the sequence $K'_l$,
\begin{equation}
\begin{aligned}
&0<K'_1<1,\\
&K'_l=\Gamma(K'_{l-1}),
\end{aligned}
\end{equation}
is also convergent \cite{huang2020deep}. We  have $\lim_{l\to \infty}K'_l=\lim_{l\to \infty}K_l=1$. The aim of us is to show $\sum_{l=1}^{L} \sqrt{(1-K_l)}\leq \cO(\log L)$.

Let $e_l=1-K_l$. Claim \ref{cl2} and \ref{cl1} below show that $e_l\sim \frac{1}{l^2}$ and
\begin{equation}
\begin{aligned}
\text{The constant term in }&\{\prod_{l=1}^{L}\Sigma(\langle h_{l-1}, h'_{l-1} \rangle/Q_{l}+ \frac{1}{L^3}\langle X_{l},X'_{l} \rangle/Q_{l})\}\geq \Omega(1/L^b).
\end{aligned}
\end{equation}
and  in this case, $b=3+\frac{\log^2L}{L}\leq 4$.
Then (\ref{eb}) follows.
\QEDA

\begin{cla}\label{c51}
Let $D$ and $D'$ be diagonal matrix satisfying
\begin{equation}
\begin{aligned}
&(D)_{k,k}=\mathbbm{1}\{\bm{W}Y+\bm{A}X>0\},\\
&(D')_{k,k}=\mathbbm{1}\{\bm{W}Y'+\bm{A}X'>0\}.
\end{aligned}
\end{equation}
If $\langle Y, Z\rangle, \langle Y, Z'\rangle =0$,
\begin{equation}\label{cle}
\begin{aligned}
\mathbb{E}_{\bm{W},\bm{A}} \langle D\bm{W}Z, D'\bm{W}Z' \rangle = &Z^TZ'\cdot \mathbb{E}_{w\sim N(0,I_m),a\sim N(0,\frac{1}{L^3}I_d)}  \langle \phi'([w,a]^T[Y,X]), \phi'([w,a]^T[Y',X'])\\
=&Z^T Z'\cdot\Sigma(\{Y^TY'+X^TX'\}/(||Y||\cdot ||Y'||+||X||\cdot ||X'||))
\end{aligned}
\end{equation}
with $\phi'(x)=\sqrt{2}\mathbbm{1}\{x>0\}.$
\end{cla}
{\bf Proof of Claim \ref{c51}:}\\
In fact,
\begin{equation}\label{cle2}
\begin{aligned}
\mathbb{E}_{\bm{W},\bm{A}} \langle D\bm{W}Z, D'\bm{W}Z' \rangle= &\langle Z',\nabla_{Y'}\langle Z, \nabla_{Y} \mathbb{E}_{w\sim N(0,I_m),a\sim N(0,\frac{1}{L^3}I_d)} \\
 &\langle \phi([w,a]^T[Y,X]), \phi([w,a]^T[Y',X']) \rangle \rangle .
\end{aligned}
\end{equation}
with $\phi(x)=\sqrt{2}\max(0,x)$. Then (\ref{cle}) is clearly a corollary of (\ref{cle2}) and $\langle Y, Z\rangle, \langle Y, Z'\rangle =0$.

\begin{cla}\label{cl1}
Supposing $K_l\sim cos[\pi(1-(\frac{l}{l+1})^{b})]+\xi_l$, $\sum_{l=l_1}^L\sqrt{\xi_l}\leq \cO(1)$, $b>0$,
\begin{equation}
\prod_{l=1}^{L}\frac{\pi-arccos(K_l)}{\pi}\geq \Omega( exp(-b\log L ))\geq \Omega (L^{-b})
\end{equation}
\end{cla}
{\bf Proof:}We use the inequality,
\begin{equation}
\prod_{l=1}^{L}(1-\frac{b}{l}-\Omega(\sqrt{\xi_l}))\geq \Omega(exp(-\sum_{l=1}^L\frac{b}{l}-\Omega(\sqrt{\xi_l}))).
\end{equation}
Meanwhile, for harmonic series,
\begin{equation}
\sum_{l=1}^L\frac{b}{l}=b\log L+b\gamma+O(1/L^2)
\end{equation}
where $\gamma \approx 0.57721 $ is the Euler- Mascheroni constant. Thus the claim follows.
\QEDA

\begin{cla}\label{cl2}
Let $e_l$ satisfy
\begin{equation}
\begin{aligned}
&e_l=\frac{Q_{l-1}}{Q_{l}}e_{l-1}+\frac{Q_{l}-Q_{l-1}}{Q_{l}}\\
 &- \frac{\sqrt{1-(1-\frac{Q_{l-1}}{Q_{l}}e_{l-1}-\frac{Q_{l}-Q_{l-1}}{Q_{l}})^2}-arccos(\frac{Q_{l-1}}{Q_{l}}e_{l-1}+\frac{Q_{l}-Q_{l-1}}{Q_{l}})(1-\frac{Q_{l-1}}{Q_{l}}e_{l-1}-\frac{Q_{l}-Q_{l-1}}{Q_{l}})}{\pi}.
\end{aligned}
\end{equation}
For $l,L$ large enough , we have $e_l\leq  1-cos[\pi(1-(\frac{l}{l+1})^{3+\frac{\log^2L}{L}})]+\xi_l$ and $\sum_{l=l_1}^L\sqrt{\xi_l}\leq \cO(1)$.
\end{cla}
Before proving this claim, we cite the following lemma in the proof of Lemma 15 in \cite{huang2020deep}:
\begin{lem}\label{lr}
Let
\begin{equation}
z_l= 1-cos[\pi(1-(\frac{l}{l+1})^{3+\frac{\log^2L}{L}})].
\end{equation}
\begin{equation}\label{er}
z_{l}\geq z_{l-1}- \frac{\sqrt{1-(1-z_{l-1})^2}-arccos(z_{l-1})(1-z_{l-1})}{\pi}+\frac{3\pi^2\log^2L}{l^3L}+\frac{20\pi^2}{2l^4}
\end{equation}
\end{lem}
{\bf Proof of Claim \ref{cl2}:}
Firstly, note that from the assumption of $||X_l||$, we have
$$\frac{Q_{l}-Q_{l-1}}{Q_{l}}\leq \cO(\frac{1}{L^3}).$$
We will show there exits $q_l=z_l+\xi_l$ such that
\begin{equation}\label{fg}
\begin{aligned}
q_l\geq &\frac{Q_{l-1}}{Q_{l}}q_{l-1}+\frac{Q_{l}-Q_{l-1}}{Q_{l}}\\ &-\frac{\sqrt{1-(1-\frac{Q_{l-1}}{Q_{l}}q_{l-1}-\frac{Q_{l}-Q_{l-1}}{Q_{l}})^2}-arccos(\frac{Q_{l-1}}{Q_{l}}q_{l-1}+\frac{Q_{l}-Q_{l-1}}{Q_{l}})(1-\frac{Q_{l-1}}{Q_{l}}q_{l-1}-\frac{Q_{l}-Q_{l-1}}{Q_{l}})}{\pi}.
\end{aligned}
\end{equation}
Then $e_l\leq q_l$. The theorem follows.

Let
\begin{equation}
\begin{aligned}
&\frac{Q_{l}-Q_{l-1}}{Q_{l}}=\epsilon_lz_{l-1},\\
&q_{l-1}=(1+\theta_l)z_{l-1},\\
&(1+\theta_{l+1})=\frac{Q_l-Q_{l-1}}{Q_l}(1+\theta_l)+\epsilon_l,\\
&\theta_{l_0+1}=0.
\end{aligned}
\end{equation}
Since $z_l<1$, $\theta_{l}>0$. And
$$\frac{Q_{l-1}}{Q_{l}}q_{l-1}+\frac{Q_{l}-Q_{l-1}}{Q_{l}}=(\frac{Q_l-Q_{l-1}}{Q_l}(1+\theta_l)+\epsilon_l)z_{l-1}=(1+\theta_{l+1})z_{l-1}.$$
Using Lemma \ref{lr},
since $\frac{\sqrt{1-z^2}-arccos(z)(1-z)}{\pi}\sim O(z^{3/2}),$ we claim that
\begin{equation}
\begin{aligned}
(1+\theta_{l+1})z_{l}\geq &(1+\theta_{l+1})z_{l-1}\\
&- \frac{\sqrt{1-(1-(1+\theta_{l})z_{l-1})^2}-arccos((1+\theta_{l})z_{l-1})(1-(1+\theta_{l})z_{l-1})}{\pi}
\end{aligned}
\end{equation}
This is because $\theta_{l}>0$, $(1+\theta_{l})^{3/2} \geq (1+\theta_{l+1})$. Then we have
\begin{equation}
\begin{aligned}
-&\frac{\sqrt{1-(1-\frac{Q_{l-1}}{Q_{l}}q_{l-1}-\frac{Q_{l}-Q_{l-1}}{Q_{l}})^2}-arccos(\frac{Q_{l-1}}{Q_{l}}q_{l-1}+\frac{Q_{l}-Q_{l-1}}{Q_{l}})(1-\frac{Q_{l-1}}{Q_{l}}q_{l-1}-\frac{Q_{l}-Q_{l-1}}{Q_{l}})}{\pi}\\
&\leq (1+\theta_{l+1})(z_l-z_{l-1}).
\end{aligned}
\end{equation}
Therefore,
\begin{equation}
\begin{aligned}
&\frac{Q_{l-1}}{Q_{l}}q_{l-1}+\frac{Q_{l}-Q_{l-1}}{Q_{l}}-\\ &\frac{\sqrt{1-(1-\frac{Q_{l-1}}{Q_{l}}q_{l-1}-\frac{Q_{l}-Q_{l-1}}{Q_{l}})^2}-arccos(\frac{Q_{l-1}}{Q_{l}}q_{l-1}+\frac{Q_{l}-Q_{l-1}}{Q_{l}})(1-\frac{Q_{l-1}}{Q_{l}}q_{l-1}-\frac{Q_{l}-Q_{l-1}}{Q_{l}})}{\pi}\\
&\leq [\frac{Q_l-Q_{l-1}}{Q_l}](1+\theta_l)z_{l-1}+\epsilon_lz_{l-1}+(1+\theta_{l+1})z_l-(1+\theta_{l+1})z_{l-1}\\
&= [\frac{Q_l-Q_{l-1}}{Q_l}(1+\theta_l)+\epsilon_l-(1+\theta_{l+1})]z_{l-1}+[1+\theta_{l+1}]z_l\\
&=[1+\theta_{l+1}]z_l\\
&=q_l.
\end{aligned}
\end{equation}
Since
\begin{equation}
(1+\theta_{l+1})=\frac{Q_l-Q_{l-1}}{Q_l}(1+\theta_l)+\epsilon_l,
\end{equation}
we can write
\begin{equation}
(1+\theta_{l+1})=1+\sum_{l'=l_0}^{l}\prod_{j=l'}^{l}\frac{Q_j-Q_{j-1}}{Q_j}\epsilon_{l'}.
\end{equation}
Then
\begin{equation}
\begin{aligned}
(1+\theta_{l+1})\leq  1+\cO(\sum_{l'=l_0}^{l}\epsilon_{l'})
\end{aligned}
\end{equation}

\begin{equation}
\begin{aligned}
q_l&=(1+\theta_{l+1})z_{l}\leq z_l+ \cO(\sum_{l'=l_0}^{l}\epsilon_{l'}z_l)\\
&\leq z_l+\cO(\sum_{l'=l_0}^{l}\frac{Q_{l'}-Q_{l'-1}}{Q_{l'}}\frac{(l')^2}{l^2})\\
&\leq z_l+\cO(\frac{l}{L^3})
\end{aligned}
\end{equation}

Since
\begin{equation}
\sum_{l=1}^L \sqrt{\frac{l}{L^3}}\leq \cO(1),
\end{equation}
 the theorem follows.
\QEDA

\section{Complexity of Functions: Proof of Theorem \ref{mc}.}
In this section, we give the detailed proof of Theorem \ref{mc}.

\begin{lem}\label{1ch}
Let
\begin{equation}
\Sigma(x)=\frac{1}{2}+\frac{arcsin(x)}{\pi}.
\end{equation}
If $||Z_i||,||Z_j||\leq \cO(1)$, $\mu>1$,
\begin{equation}
\Sigma(\{\mu+\frac{1}{L^3}Z_{i}^TZ_{j}\}/Q_{l})\succeq  \Omega( \frac{1}{L^3}) \cdot (\frac{1}{\cO(L)})^{k}\cdot \frac{1}{k^2} (Z_{i}^TZ_{j})^{k}/(||Z_{i}||\cdot Z_{j}||)^{k}
\end{equation}
\end{lem}
{\bf Proof:}
From the Taylor formula, for all $p\in\mathbb{N}$,
\begin{equation}
\begin{aligned}
\Sigma(Z_{i}^TZ_{j})\succeq \sum_{p=1}^\infty \frac{(Z_{i}^TZ_{j})^{2p-1}}{2\pi (2p-1)^2}.
\end{aligned}
\end{equation}
And
\begin{equation}
\begin{aligned}
\Sigma([\mu+\frac{1}{L^3}Z_{i}^TZ_{j}]/Q_l)\succeq \sum_{p=1}^\infty \frac{(\mu+\frac{1}{L^3}Z_{i}^TZ_{j} )^{2p-1}}{2\pi (2p-1)^2\cdot Q_l^{2p-1}}.
\end{aligned}
\end{equation}
For any $k\in \mathbb{N}$, the coefficient of $[Z_{i}^TZ_{j}/L^3]^{k}/$ in $\Sigma([\mu_0+\frac{1}{L^3}Z_{i}^TZ_{j} ]/Q_l)$ will be larger than $\frac{a_{k}}{Q_l^{k}}$ with
\begin{equation}
a_{k}=\sum_{2p-1>k}^\infty \frac{1}{2\pi (2p-1)^2}\cdot (\frac{\mu}{Q_l})^{2p-1-k}\cdot \frac{2p-1\cdot (2p-2) \cdot ... \cdot (2p-k)}{k!}
\end{equation}
Consider
\begin{equation}
b_{k}=\sum_{2p-1>k}^\infty  (\frac{\mu}{Q_l})^{(2p-1-k-2)}\cdot \frac{2p-1\cdot (2p-2) \cdot ... \cdot (2p-k+2))}{k!}
\end{equation}
$b_{k}=\Omega((\frac{\mu}{Q_l})^{2})\cdot a_k$.
Let
\begin{equation}
f(x)=\frac{1}{1-x^2}.
\end{equation}
Then
\begin{equation}
\begin{aligned}
&b_{k}\geq \Omega(|f^{(k-2)}(\frac{\mu}{Q_l})|\cdot \frac{1}{k!}\\
&=\frac{(k-2)!}{2\cdot k!}[\frac{1}{(1-\frac{\mu}{Q_l})^{k-1}}+\frac{(-1)^{k-2}}{(1+\frac{\mu}{Q_l})^{k-1}}\\
&\geq \Omega(\frac{1}{ (k-2)\cdot (k-3)}\cdot \frac{Q_l^{k-1}}{(Q_l-\mu)^{k-1}})
\end{aligned}
\end{equation}
Thus  the coefficient of $(Z_{i}^TZ_{j}/L^3)^{k}$ in $\Sigma([\mu_0+Z_{i}^TZ_{j} ]/Q_l)$ will be larger than
$$ \Omega(\frac{1}{ (k-2)\cdot (k-3)}\cdot \frac{Q_l^{-1}}{(Q_l-\mu)^{k-1}} \frac{Q_l^2}{\mu^2})\geq \Omega(Q_l-\mu)\cdot  \frac{1}{(Q_l-\mu)^{k}\cdot k^2}.$$
Since
\begin{equation}
\begin{aligned}
&0<C_1\leq ||X_{l,i}||^2, ||X_{l,j}||^2\leq  C_2,\\
& ||Z_i||^2/\sum_{l=1} ^L ||X_{i, l}||^2\sim \frac{1}{L},\\
& ||Z_j||^2/\sum_{l=1} ^L ||X_{j, l}||^2\sim \frac{1}{L}
\end{aligned}
\end{equation}
and
\begin{equation}
Q_l=\sqrt{(1+\frac{1}{L^3}\sum_{k=1}^l||X_{i,k}||^2)\cdot(1+\frac{1}{L^3}\sum_{k=1}^l||X_{j,k}||^2) }.
\end{equation}
We have
\begin{equation}
\Sigma(\{\mu+\frac{1}{L^3}Z_{i}^TZ_{j}\}/Q_{l})\succeq  \Omega( \frac{1}{L^3}) \cdot (\frac{1}{\cO(L)})^{k}\cdot \frac{1}{k^2} (Z_{i}^TZ_{j})^{k}/(||Z_{i}||\cdot ||Z_{j}||)^{k}.
\end{equation}
The claim follows.
\QEDA

Using this lemma, note that  we can write $[K^{l-1}_{i,j}+\frac{1}{L^3}X_{i,l}^TX_{j,l}]/Q_l = \frac{\mu+\frac{1}{L^3}X_{i,l}^TX_{j,l} +T_{i,j}}{Q_l}$ with $T_{i,j} \succeq 0$ where $\mu= 1$ is the constant term in $K^{l-1}_{i,j}$ and
$$[K^{l-1}_{i,j}+\frac{1}{L^3}X_{i,l}^TX_{j,l}+ T_{i,j}]/Q_l \succeq \frac{\mu+\frac{1}{L^3}X_{i,l}^TX_{j,l}}{Q_l}.$$
We have the following lemma:
\begin{lem}\label{2ch}
Under the condition of Lemma \ref{io}, for any $k\in \mathbb{N}$,
\begin{equation}
\Sigma(\{K^{l}_{i,j}+\frac{1}{L^3}X_{i,l}^TX_{j,l}\}/Q_{l})\succeq  \Omega( \frac{1}{L^3}) \cdot (\frac{1}{\cO(L)})^{k}\cdot \frac{1}{k^2} (X_{i,l}^TX_{j,l})^{k}/(||X_{i,l}||\cdot X_{j,l}||)^{k}.
\end{equation}
\end{lem}

Now we can prove Theorem \ref{mc}.
\begin{theo}\label{tlast}
Assume there is $\delta\in [0,e^{-1}]$. Let $n$ samples in $\dataset$  be $\{x_i,y_i\}_{i=1}^n$. $\widetilde{y}=[F^*(x_1),F^*(x_2),...F^*(x_n)]^T$. $F^*$ is a function belonging to the concept class (\ref{cop}) or (\ref{cop2}) such that $y_i\cdot F^*(x_i)\geq 1$ for all $i$.  There exits matrix $\bm{H}^\infty$ satisfying:
\begin{equation}
\bm{H} +\bm{\epsilon}^T\bm{\epsilon} \succeq \bm{H}^\infty \text{ with } ||\bm{\epsilon}||_F\leq 0.01/ \cO(\mathscr{C}^*)
\end{equation}
and
\begin{equation}
\sqrt{ \widetilde{y}^T (\bm{H}^\infty)^{-1}\widetilde{y}}\leq \cO(\mathscr{C}^*).
\end{equation}
\end{theo}
{\bf Proof:}

Firstly, using the forward and backward correlation Theorem \ref{tt5} and Theorem \ref{tt6},
\begin{equation}
\frac{1}{m}\langle \text{Back}_l(x_i)\cdot D_l,\text{Back}_{l}(x_j)\cdot D_{l}' \rangle \succeq \frac{1}{\cO(L^4)}\Sigma( \{\frac{1}{L^3}\langle X_{i,l}, X_{j,l} \rangle +K^{l-1}_{i,j}\}/Q_l)\pm \cO(\frac{L^4\log^4 m}{m^{1/4}})
\end{equation}
and
\begin{equation}
|\frac{1}{m}\langle \text{Back}_l(x_i)\cdot D_l,\text{Back}_{l'}(x_j)\cdot D_{l'}' \rangle| \leq \cO(\frac{L^4\log^4 m}{m^{1/4}}).
\end{equation}
for $l\neq l'$.

Thus
\begin{equation}
\begin{aligned}
H_{i,j}=\frac{1}{m}\langle \nabla f(\bm{W},x_i), \nabla f(\bm{W},x_j)\rangle= &\sum_{l=1}^L\frac{1}{m} \langle, \text{Back}^T_lD_l\cdot h_l^T(x_i), \text{Back}^T_lD_l\cdot h_l^T(x_j)\rangle \\
& \pm \cO(\frac{L^6\log^4 m}{m^{1/4}})
\end{aligned}
\end{equation}

The closure of multiplication Proposition \ref{c1} for positive definite function concludes there exits semi-positive define matrix $\bm{M}$
\begin{equation}
H_{i,j} +M_{i,j} \succeq \frac{1}{\cO(L^4)}\Sigma( \{\frac{1}{L^3}\langle X_{i,l}, X_{j,l} \rangle +K^{l-1}_{i,j}\}/Q_l).
\end{equation}
with $M_{i,j}\leq \cO(\frac {L^6\log^2 m}{m^{1/4}})$. Then $||\bm{M}||_F \leq n^2\frac{L^6\log^2m}{m^{1/4}}$.   $\bm {M} $ is semi-positive define, therefore there exits $\bm{\epsilon}^T \bm{\epsilon }=\bm{M}$,  $||\bm {\epsilon} ||_F \leq 0.01/\mathscr{C}^*$ by SVD and reshaping since  $m>poly(n,\mathscr{C}^*)$. Meanwhile  let $$\bm{G}=m^{-1/2}\cdot (vec[\nabla f(\bm{W}^0,x_1)], vec[\nabla f(\bm{W}^0,x_2)],... vec[\nabla f(\bm{W}^0,x_n)])\in \mathbb{R}^{m^2\times n}.$$ Since $m^2\gg n$,  we can set $\bm{\epsilon}$ satisfying  $\bm{\epsilon}^T \bm{G} =0$ without changing $\bm{\epsilon^T\epsilon}$.

For  a function $\psi(\beta^T_{l,r}X_l/||X_{l}||)=\sum_{p=1}^\infty c_p (\beta^T_{l,r}X_l/||X_{l}||)^{p}$ with $||\beta_{l,r}||\leq 1$, let
\begin{equation}
y_p=[c_p (\beta^T_{l,r}X_{1,l}/||X_{1,l}|)^{p},...c_p(\beta^T_{l,r}X_{n,l}/||X_{n,l}||)^{p}]\in \mathbb{R}^n.
\end{equation}
Using Proposition \ref{claa}, if  $H^\infty_{i,j}\succeq \xi_p(X_{l,i}^T X_{l,j}/(||X_{l,i}||\cdot ||X_{l,j}||))^{p}$,
 $$y_p^T(\bm{H}^\infty)^{-1}y_p\leq \frac{c^2_p||\beta^T_{l,r}||^{2p}}{\xi_p}.$$

In our case, from Lemma \ref{2ch},
\begin{equation}
\xi_p=\Omega( \frac{1}{L^4})\cdot \Omega( \frac{1}{L^3}) \cdot (\frac{1}{\cO(L)})^{p}\cdot \frac{1}{p^2}
\end{equation}
Note that $$ y\overset{\text def}{=}\sum_{p=1}^\infty y_p=[\psi(\beta^T_{l,r}X_{1,l}/||X_{1,l}||),...\psi(\beta^T_{l,r}X_{n,l}/||X_{n,l}||)]\in \mathbb{R}^n.$$
We have
\begin{equation}\label{elast}
\sqrt{y^T(\bm{H}^\infty)^{-1}y}\leq \sum_p \sqrt{y^T_p(\bm{H}^\infty)^{-1}y_p}\leq \sum_{p=1}^\infty \frac{c_p||\beta^T_{l,r}||^{p}}{\xi_p}.
\end{equation}
In our case ,
\begin{equation}
\frac{1}{\sqrt{\xi_p}}\leq O( L^{3.5}) \cdot (\cO(\sqrt{L}))^{p}\cdot p.
\end{equation}
 We have
\begin{equation}\label{comp}
\sqrt{\widetilde{y}^T (\bm{H}^\infty)^{-1}\widetilde{y}}\leq \cO(\mathscr{C}^*)
\end{equation}
for Additive Concept Class (\ref{cop}).

For N-variables Concept Class (\ref{cop2})
$$F^*(x)=\sum_{r}\psi_{r}(\langle \beta_{r}, [X_{l_1},...,X_{l_{N}}]\rangle/\sqrt{N}\max||X_{l_n}||).$$

We rewrite it as
$$F^*(x)=\sum_{r}\psi_{r}(\langle \beta_{r}, [X_{l_{max}},...,X_{l_{max}-N'}]\rangle/\sqrt{N}\max||X_{l_n}||$$ $$l_{max}=\max(l_1,..,l_N), N'=\max(l_1,..,l_N)-\min(l_1,..,l_N).$$

Finally we prove that
$\sqrt{\widetilde{y}^T (\bm{H}^\infty)^{-1}\widetilde{y}}\leq \cO(L^4\sum_r\mathscr{C}_N(\psi_{r},\cO(\sqrt{L}))$.

Based on the structure of $H^\infty$, we have
$$H^\infty_{i,j}\succeq \frac{1}{\cO(L^4)}\cdot \Sigma(\{K^{l_1}_{i,j}+\frac{1}{L^3}X_{i,l_1}^TX_{j,l_1}\}/Q_{l_1})\cdot... \cdot \Sigma(\{K^{l_N}_{i,j}+\frac{1}{L^3}X_{i,l_N}^TX_{j,l_N}\}/Q_{l_N}).$$
Then we have the follow claim
\begin{cla}\label{cf}
For any N terms $X_{i,r_1}^TX_{j,r_1}, X_{i,r_2}^TX_{j,r_2}...,X_{i,r_N}^TX_{j,r_N}$, $r_{max}=\max(r_1,...r_N)$, we have
\begin{equation}
\begin{aligned}
H^\infty_{i,j}&\succeq \frac{1}{C_1^N L^4 \cdot L^{2N} \cdot C_{N,p}\cdot(p/N)^{2N}}\\
&\cdot (X_{i,r_1}^TX_{j,r_1}/||X_{i,r_1}||\cdot ||X_{j,r_1}|| +X_{i,r_2}^TX_{j,r_2}/||X_{i,r_2}||\cdot ||X_{j,r_2}||...\\
& + X_{i,r_N}^TX_{j,r_N}/||X_{i,r_N}||\cdot ||X_{j,r_N}||)^{p}\\
&\succeq \frac{1}{C_1^N L^4 \cdot L^{2N} \cdot C_{N,p}\cdot(p/N)^{2N}}\\
&\cdot (X_{i,r_1}^TX_{j,r_1} +X_{i,r_2}^TX_{j,r_2}... + X_{i,r_N}^TX_{j,r_N})^p/(N\cdot \max_n(||X_{i,r_n}||)\cdot \max_n(||X_{j,r_n}||)) ^{p}
\end{aligned}
\end{equation}
where $C_1$ is a large constant.
\end{cla}
which can be deduced from the following facts:
\begin{enumerate}[(a)]
\item For $k\in\mathbb{N}$, $\Sigma(\{K^{l}_{i,j}+\frac{1}{L^3}X_{i,l}^TX_{j,l}\}/Q_{l})\succeq  \Omega( \frac{1}{L^3}) \cdot (\frac{1}{\cO(L)})^{k}\cdot \frac{1}{k^2} (X_{i,l}^TX_{j,l})^{k}/(||X_{i,l}||\cdot X_{j,l}||)^{k}.$
\item For any $n$ integers $n_1,n_2,...n_N$, with $n_1+n_2+..+n_N=p$,  $C_{N,p}\geq \frac{p!}{n_1!n_2!...n_N!}$ and the largest coefficient of monomial in $(x_1+x_2+,..+x_N)^{2p-1}$ is less than $C_{N,p}$.
\item For any $n$ integers $n_1,n_2,...n_N$, with $n_1+n_2+..+n_N=p$, $(p/N)^{2N}\geq n^2_1\cdot...\cdot n^2_N)$.
\end{enumerate}

(b) and (c) are trivial. (a) is from Lemma \ref{2ch}.

Combing these results, polynomial theorem  and using a similar argument as (\ref{comp}), we have
$$\sqrt{\widetilde{y}^T (\bm{H}^\infty)^{-1}\widetilde{y}}\leq L^2\cO(1+\sum_{p=1}^\infty  L^{1.5N} C_1^N\cdot \sqrt{C_{N,p}}\cdot (p/N)^N (\cO(\sqrt{L}))^{p} \cdot |c_{p}|.$$

Thus $\sqrt{\widetilde{y}^T (\bm{H}^\infty)^{-1}\widetilde{y}}\leq \cO(L^2\sum_r\mathscr{C}_N(\psi_{r},1))$.

Finally we prove $$\sqrt{\widetilde{y}^T (\bm{H}^\infty)^{-1}\widetilde{y}}\leq \cO(L^3\sum_r\mathscr{C}(\psi_{r},2^{l_0}\cO(\sqrt{L}))).$$
Consider
\begin{equation}
\begin{aligned}
&K^1_{i,j}=Q_l\cdot \sum_{r=0}^\infty \mu^2_r (1+\frac{1}{L^3}X_{i,1}^TX_{j,1}/Q_l)^r,\\
&K^{l}_{i,j}=Q_l\sum_{r=0}^\infty \mu^2_r (\{\frac{1}{L^3}X_{i,l}^TX_{j,l}+ K^{l-1}_{i,j}\}/Q_l)^r.
\end{aligned}
\end{equation}
with  $\mu_r=\frac{1}{\sqrt{2\pi}}\int_0^\infty \sqrt{2}xh_r(x)e^{-\frac{x^2}{2}}dx$, $h_r(x)=\frac{1}{\sqrt{r!}}(-1)^re^{\frac{x^2}{2}}\frac{d^r}{dx^r}e^{-\frac{x^2}{2}}.$
We can rewrite this equation as:
\begin{equation}
\begin{aligned}
&\overline{K}_1=\sum_{r=0}^\infty \mu^2_r (1+\frac{1}{L^3}X_{i,1}^TX_{j,1}/Q_1)^r=\Gamma([1+\frac{1}{L^3}X_{i,1}^TX_{j,1}]/Q_1) ,\\
&\overline{K}_l=\Gamma(\overline K_{l-1}\cdot Q_{l-1}/Q_l),\\
&\Gamma(x)=x+\frac{\sqrt{1-x^2}-arccos(x)x}{\pi}.
\end{aligned}
\end{equation}
and
\begin{equation}\label{ev}
\begin{aligned}
K^{l}_{i,j}=&\overline{K}_l\cdot Q_l\\
=&Q_l\cdot \underbrace{\Gamma \frac{Q_{l-1}}{Q_{l}}\circ...\circ\Gamma \{ \frac{1}{Q_1}\cdot (1+\frac{1}{L^3}X_{i,1}^TX_{j,1})\}}_{l\ times}.
\end{aligned}
\end{equation}
Using the fact
\begin{equation}
Q_l \cdot \prod_{k=k_0}^{l}\frac{Q_{l-1}}{Q_l}=Q_{k_0-1},
\end{equation}
and $$\nabla_x f_l \circ f_{l-1}...\circ f_1(x)= f_l'\circ f'_{l-1}...\circ f'_1(x),$$
The linear part in $K^{l}_{i,j}$ is $\sum_{r=0}^{l-1} \mu_1^{2l-2r} \frac{1}{L^3}X_{i,r}^TX_{j,r}$. Thus
\begin{equation}
\begin{aligned}
&K^{l}_{i,j}+\frac{1}{L^3}X_{i,l}^TX_{j,l}\\
&\succeq \frac{1}{L^3}X_{i,l}^TX_{j,l}+ \sum_{r=0}^{l-1} \mu_1^{2l-2r} \frac{1}{L^3}X_{i,r}^TX_{j,r}\\
&\succeq \mu_1^{2l} \sum_{r=1}^l  \frac{1}{L^3}X_{i,r}^TX_{j,r}
\end{aligned}
\end{equation}
with $\mu_1^2=\frac{1}{2}$.

$||\mu_1^{2l} \sum_{r=1}^l X_{i,r}^TX_{j,r}||\leq \mu_1^{2l} \cdot l\leq \cO(1)$. Then from Lemma \ref{1ch}, we have
\begin{equation}
H^{\infty}_{i,j}\succeq\Omega( \frac{1}{L^7}) \cdot (\frac{1}{\cO(L)})^{k}\cdot \frac{1}{k^2} (Z_{i}^TZ_{j})^{k}/(||Z_{i}||\cdot ||Z_{j}||\cdot 2^{l})^{k}
\end{equation}
with $Z_{i}^TZ_{j}= \sum_{r=1}^l  \mu_1^{2r}X_{i,r}^TX_{j,r}$ and $||Z_{i}||^2=\sum_r  \mu_1^{2r}||X_{i,r}||^2$

Therefore
$$\sqrt{\widetilde{y}^T (\bm{H}^\infty)^{-1}\widetilde{y}}\leq \cO(L^{3.5}\sum_r\mathscr{C}(\psi_{r},2^{l_0}\cO(\sqrt{L})))$$

The theorem follows.
\QEDA
\begin{rem}\label{g1}
Based on the previous results, we can generalize the results to the loss with the form: $$\frac{1}{n}\sum_{i=1}^n\sum_{l=1}^L\ell(y_i\cdot f_l(\bm{W},x_i))$$ with $f_l(\bm{W},x)=\bm{B}^Th_l(x)$ to show for $\bm{H}^l_{i,j}=\langle \nabla f_l(\bm{W},x_i), \nabla f_l(\bm{W},x_j)\rangle$, there exits
 \begin{equation}
\bm{H}^l +\bm{\epsilon}^T\bm{\epsilon} \succeq (\bm{H}^l)^\infty \text{ with } ||\bm{\epsilon}||_F\leq 0.01/ \sqrt{ \widetilde{y}^T ((\bm{H}^l)^\infty)^{-1}\widetilde{y}}
\end{equation}

In fact we have following two generalization results of previous results which are in fact already contained in the proof.

Generalization of Lemma \ref{l3}:

Let $g_l=\phi_1(\bm{W}g_{l-1}),   h_l(x_1)=\phi_2(\bm{W}h_{l-1}(x_l)+\bm{A}X_l)$. $\widetilde{g}_l=\phi_1(\widetilde{\bm{W}}\widetilde{g}_{l-1})$ and $\widetilde{h}_l(x_1)=\phi_1(\widetilde{\bm{W}}\widetilde{h}_{l-1}(x_l)+\widetilde{\bm{A}}X_l)$ are defined by  $\widetilde{\bm{W}}, \widetilde{\bm{A}}$. $\widetilde{\bm{W}}, \widetilde{\bm{A}}$ and $\bm{W}, \bm{A}$ are i.i.d. Then for any $0<l,l'<L$, with probability at least $1-L^2exp(-\Omega(\log^2m))$,
\begin{equation}
|g_{l}^Th_{l'}(x')-  {\overline{g}_{l}}^{T}\overline{h}_{l'}(x')|\leq \cO(L^2\log^2m/m)
\end{equation}
where
$$\overline{g}_l=\phi_1(\bm{W}\widetilde{g}_{l-1})$$
$$\overline{h}_{l'}(x_1)=\phi_2(\bm{W}\widetilde{h}_{l'-1}(x_l)+\bm{A}X_l)$$

Let $\phi_1(x)=x, \phi_2(x)=max(x,0)$. One corollary of this result is that  from (4.2) in \cite{rate}, there exits  $g_l$, such that  $\langle g_l, h_{l'}\rangle \geq 1/poly(L)$ when $l=l'$. Else $\langle g_l, h_{l'}\rangle =0$.

Generalization of Theorem \ref{t17}:

With probabiluty at least $1-L^2exp(-\Omega(\log^2m))$,
\begin{equation}
|\frac{1}{m}\langle BD_{l_1}W\cdots D_{l+1}WD_l,BD'_{l_2}W\cdots D'_{l+1}WD'_l \rangle| \leq \cO(\frac{L^4\log^4 m}{m^{1/4}}).
\end{equation}
if $l_1\neq l_2$.

 Then we can show there exits $w^*_a$ with $||w^*_a||\leq \mathscr{C}(F_a^*)$. for $a=1,2...L$ with

 \begin{equation}
\frac{1}{\sqrt{m}}\langle \nabla_{\widetilde{W}} f_a(\widetilde{\bm{W}},x_i), w^*_{a} \rangle=\frac{1}{m}\sum_{l}\langle BD_{a}W\cdots D_{l+1}WD_l,w^*_{a,back} \rangle\cdot \langle h_l(x_i), g_{a}\rangle=F^*_a(x_i)+\epsilon
\end{equation}
and
\begin{equation}
|\frac{1}{\sqrt{m}}\langle \nabla_{\widetilde{W}} f_{a'}(\widetilde{\bm{W}},x_i), w^*_{a} \rangle=|\frac{1}{m}\sum_{l}\langle BD_{a'}W\cdots D_{l+1}WD_l,w^*_{a,back} \rangle\cdot \langle h_l(x_i), g_{a}\rangle|\leq \epsilon
\end{equation}
when $a\neq a'$.

Here $w^*_{a,back}$ is from the SVD of matrix $$\frac{1}{m}\langle BD_{l_1}W\cdots D_{l+1}WD_l,BD'_{l_2}W\cdots D'_{l+1}WD'_l \rangle.$$
as (\ref{svd}) in the proof of Lemma \ref{lem2}.
\end{rem}

\end{document}